\documentclass{article}

\usepackage{graphicx}
\usepackage{makecell}
\usepackage[normalem]{ulem}
\usepackage[dvipsnames]{xcolor}
\usepackage{amsmath,amssymb,amsthm}
\usepackage{tikz}
\usepackage{multirow}
\usepackage{microtype}
\usetikzlibrary{arrows.meta, positioning, fit, calc,
                 decorations.pathreplacing}
\usepackage{pgfplots}
\usepackage{pgfplotstable}
\pgfplotsset{compat=1.18}
\usepackage[colorlinks=true, linkcolor=blue, citecolor=blue, urlcolor=blue]{hyperref}
\usepackage{subcaption}
\PassOptionsToPackage{numbers}{natbib}
\usepackage[preprint]{neurips_2026}
\usepackage{booktabs}
\usepackage{multirow}
\usepackage{graphicx}

\usepackage{xcolor}

\pgfplotsset{compat=1.18}
\usetikzlibrary{pgfplots.groupplots, calc} 
\usetikzlibrary{external}
\ifdefined\arxivbuild
  \tikzset{external/mode=only graphics}%
\else
  \ifdefined\pdfshellescape
    \ifnum\pdfshellescape=1\relax\else
      \tikzset{external/mode=only graphics}%
    \fi
  \fi
\fi

\ifdefined\showSeriesB\else
  \newif\ifshowSeriesB
  \showSeriesBtrue
\fi

\newcommand{\HFigure}[1]{\hyperref[#1]{Figure~\ref*{#1}}}

\newcommand{\OurAlgo}{dMX}

\theoremstyle{plain}

\theoremstyle{definition}

\theoremstyle{remark}

\title{dMX: Differentiable Mixed-Precision Assignment for Low-Precision Floating-Point Formats}

\author{%
  Giuseppe Franco \thanks{Correspondence to: giuseppe.franco@amd.com}\\
  \And
  Ian Colbert \\
  \And
  Pablo Monteagudo-Lago \\
  \And
  Felix Marty \\
  \And
  Nicholas Fraser \\
  \AND
  AMD
}

\begin{document}

\maketitle

\begin{abstract}
Quantizing large language models (LLMs) to low-precision
floating-point representations is central to efficient
deployment,
yet applying a single bit-width uniformly across all
layers is sub-optimal when considering the performance/accuracy trade-off.
This work introduces \OurAlgo,
a differentiable mixed-precision quantization framework
for learnable floating-point bit-width assignment.
In particular,
we study its application for the microscaling floating-point (MXFP)
family of data types defined by the Open Compute Project (OCP)
standard.
The per-layer bit-width assignment is formulated as a
continuous optimization problem in which each layer's
floating-point format format is parameterized by a scalar parameter,
folding the multi-variate design space into a single learnable offset.
During training this offset takes continuous values,
avoiding sudden oscillations between discrete quantization formats.
A temperature-based annealing schedule progressively
discretizes the learned offsets,
ensuring that the final configuration maps to
hardware-compatible MXFP formats without abrupt
transitions between training and inference behavior.
A target-aware regularization term steers the average
bit-width toward a user-specified budget,
serving as a coarse-grained proxy for inference cost
and balancing model quality against deployment
efficiency.
We performed experiments on different families of LLMs,
such as Llama, Qwen3,
and SmolLM2, evaluating perplexity on WikiText-2
and accuracy on four zero-shot reasoning benchmarks.
Across these settings, \OurAlgo~consistently
yields Pareto-dominating models and
improves over Kullback-Leibler (KL) divergence-based layer-selection
heuristics,
efficiently navigating trade-offs between model quality
and average bit-width.
\end{abstract}

\section{Introduction}\label{sec:introduction}

Deploying large language models (LLMs) at scale demands
substantial reductions in both memory footprint and
computational cost,
a challenge that grows with each new generation of model
architectures.
Quantization is one of the primary tools for achieving
this goal:
high-precision parameters are replaced with
lower-precision representations while aiming to preserve
model quality.
Full quantization-aware training (QAT) can recover much
of the lost accuracy in principle,
but it remains prohibitively expensive for modern
LLMs.
This practical limitation has motivated the development
of more expressive data types that can be used
post-training~\cite{frantar2022gptq,lin2024awq,
xiao2023smoothquant}.

\begin{figure}[h!]
\includegraphics[width=\textwidth ]{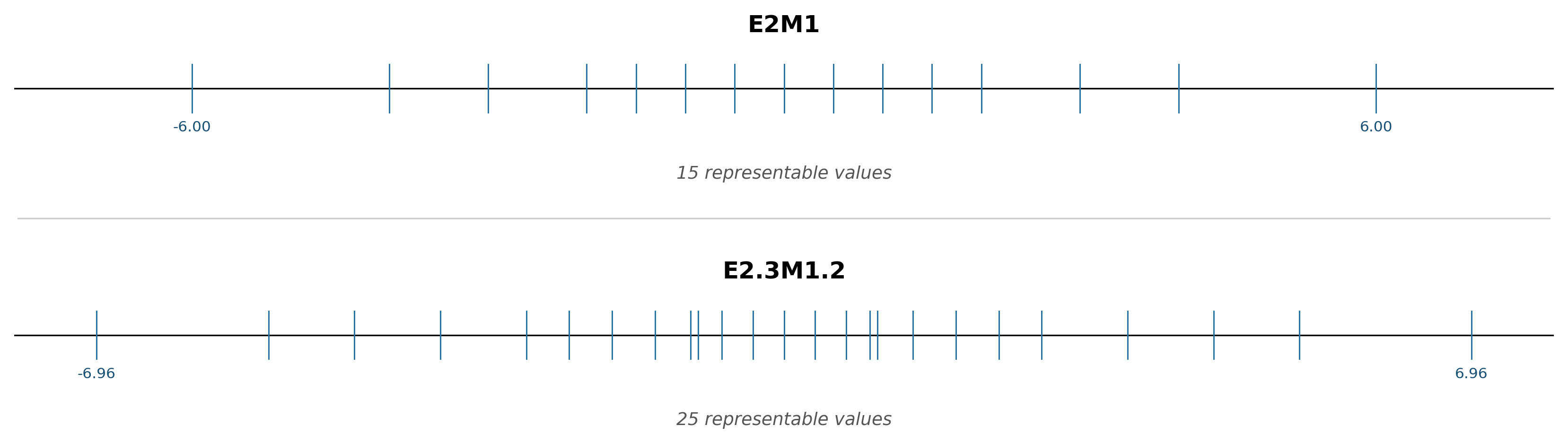}
\caption{Comparison of the quantization grids when using continuous values
for mantissa and exponent bit-widths, compared to a discrete E2M1 configuration.}
\label{fig:quantization_grids}
\end{figure}

In this context, floating-point formats such as FP8, FP6, and FP4
offer greater dynamic range than integer formats at the same
nominal bit-width~\cite{vanbaalen2023fp8vsint8}.
Per-group scaling further improves granularity relative
to per-tensor or per-channel
schemes~\cite{rouhani2023microscaling},
and dynamic activation quantization adapts to the
statistics of each input at inference time.
The microscaling (MX) standard published by the
Open Compute Project
(OCP)~\cite{ocpmx2023} formalizes a family of
block-scaled floating-point formats (i.e., MXFP8, MXFP6,
and MXFP4) that are now supported by recent accelerator
architectures.

Despite these advances,
applying a single low-precision format uniformly across
all layers still leads to substantial accuracy loss.
Deploying MXFP4 throughout an entire network,
for instance,
often degrades model quality beyond acceptable
thresholds~\cite{egiazarian2025bridginggappromiseperformance, sanjeet2026mixquant}.
This limitation motivates mixed-precision quantization,
in which different layers or operators receive different
bit-widths so that the model can achieve a more
favorable balance between accuracy and efficiency.
This approach introduces its own
challenge:
the bit-width assignment across layers is a
combinatorial optimization
problem~\cite{cohen2025gradient_free,wu2018mixed},
and estimating how quantization errors interact across
layers is
difficult~\cite{akbulut2026infoq,ranjan2025mixqsam}.
Greedy methods typically evaluate the degradation caused
by quantizing each layer in
isolation~\cite{dong2019hawq},
or they approximate cross-layer interactions only
partially through local proxies of sensitivity
~\cite{akbulut2026infoq, dong2020hawqv2}.
Their decisions depend strongly on the chosen
sensitivity metric and do not account for the nonlinear
accumulation of quantization errors through the
network.

Gradient-based methods offer a global
alternative~\cite{wu2018mixed,yang2021fracbits},
as they optimize bit-width parameters jointly through
backpropagation,
thereby capturing cross-layer dependencies that greedy
approaches
miss.
Learned bit-width representations have been studied
primarily for integer
quantization~\cite{yang2021fracbits,huang2022sdq,
yang2021bsq};
comparatively little work has explored analogous ideas
for floating-point formats.
In the floating-point setting,
optimization must contend with the structure of the
data type itself,
which involves exponent and mantissa bit-widths together
with related quantities such as the exponent bias.
Current hardware supports only a restricted set of
format combinations,
typically limited to (MX)FP8, (MX)FP6, and (MX)FP4,
with predefined choices for special-value
representation and exponent
bias~\cite{ocpmx2023}.
These constraints mean that the continuous search space
(see \HFigure{fig:quantization_grids})
must ultimately collapse to a small discrete set of
admissible configurations,
making both the parameterization of the search and the
discretization mechanism central design decisions.
Although a more thorough analysis of related works will be presented in Sec. \ref{sec:related},
to the best of our knowledge,
this is the first work
that proposes a framework for a gradient-based bit-width allocation,
focused on the MX format.

\begin{figure*}[t]
\centering
\resizebox{\textwidth}{!}{%
\begin{tikzpicture}[
    >=Stealth,
    node distance=0.4cm and 0.55cm,
    basebox/.style={
        draw, rounded corners=3pt,
        minimum height=0.85cm,
        text width=2.1cm,
        align=center,
        font=\scriptsize,
        inner sep=3pt,
    },
    inputbox/.style={
        basebox,
        fill=gray!15,
        draw=gray!60,
        text width=2.1cm,
    },
    llmlayer/.style={
        draw, rounded corners=2pt,
        minimum height=0.55cm,
        text width=1.75cm,
        align=center,
        font=\scriptsize,
        inner sep=2pt,
        fill=blue!12,
        draw=blue!60,
        line width=0.8pt,
    },
    llmframe/.style={
        draw=gray!60,
        rounded corners=4pt,
        fill=gray!5,
        inner sep=5pt,
    },
    novelbox/.style={
        basebox,
        fill=blue!12,
        draw=blue!60,
        line width=0.8pt,
    },
    stdbox/.style={
        basebox,
        fill=gray!8,
        draw=gray!50,
    },
    gradbox/.style={
        basebox,
        fill=red!10,
        draw=red!65!black,
        line width=0.8pt,
    },
    mixedbox/.style={
        basebox,
        fill=gray!5,
        draw=gray!55,
        line width=0.8pt,
        text width=3.2cm,
    },
    outputbox/.style={
        basebox,
        fill=green!12,
        draw=green!50!black,
        text width=2.1cm,
    },
    arrowstyle/.style={
        ->,
        thick,
        gray!70!black,
    },
    loopstyle/.style={
        ->,
        thick,
        blue!60!black,
        dashed,
    },
    loopstyleB/.style={
        ->,
        thick,
        red!70!black,
        densely dashdotted,
    },
    annot/.style={
        font=\tiny\itshape,
        text=gray!60!black,
        align=center,
    },
    gradannot/.style={
        font=\tiny\itshape,
        text=blue!60!black,
        align=center,
    },
    tempannot/.style={
        font=\tiny\itshape,
        text=red!70!black,
        align=center,
    },
    sublabel/.style={
        font=\tiny,
        text=black!70,
        align=center,
    },
]



\node[llmlayer] (layer1)
    {Layer 1\enspace learned $\beta_1$};
\node[llmlayer, below=3pt of layer1] (layer2)
    {Layer 2\enspace learned $\beta_2$};
\node[font=\scriptsize, text=gray!70!black, below=1pt of layer2] (layerdots)
    {$\vdots$};
\node[llmlayer, below=1pt of layerdots] (layerL)
    {Layer $L$\enspace learned $\beta_L$};

\node[llmframe,
      fit=(layer1)(layer2)(layerdots)(layerL),
      label={[font=\scriptsize\bfseries,
              text=gray!70!black]above:Pre-trained LLM}]
      (input) {};

\node[llmlayer] at (layer1) {Layer 1 \\ $\beta_1$};
\node[llmlayer] at (layer2) {Layer 2 \\ $\beta_2$};
\node[font=\scriptsize, text=gray!70!black] at (layerdots) {$\vdots$};
\node[llmlayer] at (layerL) {Layer $L$ \\ $\beta_L$};

\node[novelbox, right=of input] (fwd)
    {Forward Pass\\($\beta_i \to \hat{\beta_i}$)};

\node[stdbox, right=of fwd] (loss)
    {Task Loss\\$\mathcal{L}_{\text{task}}$};

\node[novelbox, right=of loss] (reg)
    {$\beta_i$ Regularization\\$\mathcal{R}$};

\node[mixedbox, right=of reg] (update)
    {\begin{tabular}{@{}c|c@{}}
     \textcolor{red!70!black}{Gradient} &
     \textcolor{blue!60!black}{Temperature}\\
     \textcolor{red!70!black}{Update} &
     \textcolor{blue!60!black}{Annealing}\\
     \textcolor{red!70!black}{$(\nabla_\beta, \nabla_w)$} &
     \textcolor{blue!60!black}{$T\!\uparrow$}
     \end{tabular}};

\node[outputbox, right=of update] (output)
    {Mixed-Prec.\\MXFP\\LLM};


\draw[arrowstyle] (input) -- (fwd);
\draw[arrowstyle] (fwd)   -- (loss);
\draw[arrowstyle] (loss)  -- (reg);
\draw[arrowstyle] (reg)   -- (update);
\draw[arrowstyle] (update)  -- (output);


\path let \p1=(input.south), \p2=(update.south) in
    coordinate (rowbottom) at (0, {min(\y1,\y2)});

\coordinate (tempfloor-y) at ($(rowbottom) + (0, -0.45cm)$);
\coordinate (gradfloor-y) at ($(rowbottom) + (0, -0.85cm)$);


\coordinate (loopR) at ($(update.south) + (0.82cm, 0)$);
\coordinate (loopL) at ($(fwd.south) + (-0.18cm, 0)$);

\draw[loopstyle]
    (loopR)
    -- (loopR |- gradfloor-y)      
    -| (loopL);                     


\coordinate (tempR) at ($(update.south) + (-0.82cm, 0)$);
\coordinate (loopL2) at ($(fwd.south) + (0.18cm, 0)$);

\draw[loopstyleB]
    (tempR)
    -- (tempR |- tempfloor-y)      
    -| (loopL2);                    


\node[tempannot, anchor=north]
    at ($(loopL2 |- tempfloor-y)!0.5!(tempR |- tempfloor-y)$)
    {optimize $\beta_i$ per layer};

\node[gradannot, anchor=north]
    at ($(loopL |- gradfloor-y)!0.5!(loopR |- gradfloor-y)$)
    {$T$: low$\to$high};


\node[sublabel, above=2pt of fwd.north, anchor=south,
      text width=2.1cm]
    {$\hat{\beta_i}\!=\!F(\beta_i, T)$\\No rounding;\\smooth formats};


\end{tikzpicture}%
}
\caption{\textbf{Overview of the \OurAlgo~pipeline.}
    All blue elements highlight the main contributions
    of this work.
    The pre-trained LLM (left) contains a learned
    continuous offset~$\beta_i$ for each layer~$i$,
    which parameterizes the bit-width used in that
    layer.
    During the forward pass these offsets are mapped to
    discrete format assignments
    $\hat{\beta} = F(\beta, T)$.
    A task loss and a user-defined regularization
    term~$\mathcal{R}$ on $\beta$ jointly drive the
    gradient update (red loop), while a temperature
    annealing schedule progressively sharpens the
    mapping so that every~$\beta_i$ converges to a
    hardware-compatible MXFP format (blue loop).
    }
\label{fig:method_overview}
\end{figure*}
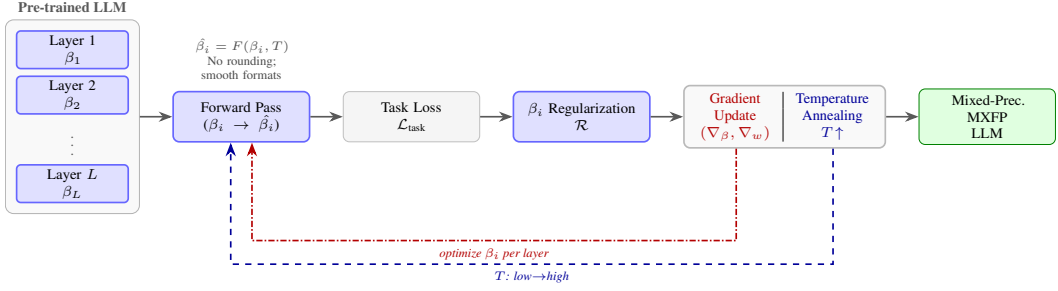

This work introduces a gradient-based framework for
learning per-layer floating-point bit-widths.
An overview of the pipeline is shown in
\HFigure{fig:method_overview}.
The main contributions are:
\begin{itemize}
    \item A differentiable bit-width parameterization
        for floating-point quantization that enables
        end-to-end gradient-based optimization of
        per-layer precision assignments.
    \item A continuous representation of the
        floating-point format of the bit-width during calibration that
        produces smoother optimization landscapes
        compared to discrete alternatives based on
        rounding and straight-through estimators
        (STEs)~\cite{bengio2013estimating}.
    \item A temperature-based annealing mechanism that
        progressively discretizes the learned
        bit-widths,
        ensuring convergence to hardware-compatible
        formats while minimizing the gap between
        calibration-time and inference-time model
        behavior.
    \item Support for targeting specific hardware
        configurations through mixed-precision
        quantization for floating-point quantization.
        We  focus our experiments on MXFP configurations,
        in particular on MXFP8/MXFP4 and
        MXFP6/MXFP4 format pairs.
    \item Compatibility with existing post-training
        quantization (PTQ) algorithms,
        enabling integration with gradient-based weight
        optimization methods and rotation-based
        preprocessing
        techniques~\cite{liu2024spinquant}.
\end{itemize}

\section{Methodology}\label{sec:methodology}

The proposed method rests on three observations,
whose combination in the floating-point setting
has not been previously explored.
The first is that bit-widths need not be integer-valued
during optimization.
The quantization equations remain well defined when the
bit-width is treated as a continuous quantity,
and no mathematical constraint forces a discrete
bit-width.

The second observation follows directly from the
deployment setting:
although continuous bit-widths are permissible during
training,
the final assignments must map to discrete,
hardware-supported formats at deployment time.
In the case of floating-point quantization,
this mapping is more restrictive than in the integer
setting.
The MX format family supports only a specific set of
configurations, each with
a fixed partition of bits between exponent and
mantissa~\cite{ocpmx2023}.

The third observation is a well-established property
of gradient-based optimization:
gradients can be propagated with respect to the
bit-width in the same way as for any other learnable
parameter,
provided that the bit-width itself is stored as a
floating-point variable.
This idea has been applied successfully to integer
bit-width
learning~\cite{yang2021fracbits,huang2022sdq};
its extension to floating-point quantization remains
largely unexplored.

These observations give rise to three design
challenges.
The first is how to parameterize the floating-point
representation so that all quantities entering the
quantizer remain differentiable with respect to the
bit-width.
The second is how to ensure that the learned bit-widths
converge to hardware-compatible values by the end of
training.
The third is how to constrain the bit-widths within
user-defined boundaries.
The following subsections address each challenge in
turn.

\subsection{Continuous floating-point bit-width
parameterization}\label{subsec:cont_param}

The standard formulation of floating-point quantization
represents a floating-point value as
\begin{equation}\label{eq:fp_value}
    x^{(\mathrm{FP})} = (-1)^{S} \times 2^{u - b}
    \times \left(1 + \sum_{i=1}^{m} M_i
    \cdot 2^{-i}\right),
\end{equation}
where $S \in \{0,1\}$ is the sign bit,
$u$ is the unsigned exponent with
$0 \leq u < 2^{e}$,
$b = 2^{e-1} - 1$ is the exponent bias,
$e$ and $m$ denote the number of exponent and mantissa
bits respectively,
and $M_i \in \{0,1\}$ is the $i$-th mantissa bit.
This formulation is adjusted accordingly in case of subnormal values.
The representable range of this format is bounded by
\begin{equation}\label{eq:fp_range}
    q_{\max}^{(\mathrm{FP})} = (2 - 2^{-m})
    \cdot 2^{\,2^{e} - b - 1},
    \qquad
    q_{\min}^{(\mathrm{FP})}
    = -q_{\max}^{(\mathrm{FP})}.
\end{equation}
Given a tensor $\mathbf{X}$ with elements
$x \in \mathbf{X}$ and a coarse scale
$s = t \,/\, q_{\max}^{(\mathrm{FP})}$ where
$t = \max(|\mathbf{X}|)$,
each quantized element $x_q$ is obtained by scaling,
rounding to the nearest representable value,
and clipping:
\begin{equation}\label{eq:fp_quant}
    x_q
    = \operatorname{clip}\!\left(
        \mathit{ss}(\tilde{x}) \left\lfloor
            \frac{\tilde{x}}{\mathit{ss}(\tilde{x})}
        \right\rceil;\;
        q_{\min}^{(\mathrm{FP})},\,
        q_{\max}^{(\mathrm{FP})}
    \right),
\end{equation}
where $\tilde{x} = x/s$ is the magnitude-scaled element
and $\lfloor \cdot \rceil$ denotes rounding to the nearest
integer.
The fine-grained power-of-two scale $\mathit{ss}(\tilde{x})$
is derived from each element's magnitude via
\begin{align}
    e_{\min} &= 1 - b - m, \label{eq:fp_emin}\\
    \eta(\tilde{x}) &= \max\!\Bigl(
        \bigl\lfloor \log_2(|\tilde{x}|) \bigr\rfloor - m,\;
        e_{\min}\Bigr), \label{eq:fp_eta}\\
    \mathit{ss}(\tilde{x}) &= 2^{\,\eta(\tilde{x})}, \label{eq:fp_ss}
\end{align}
where $e_{\min}$ is the subnormal exponent floor and
$\eta(\tilde{x})$ is the exponent of the unit in the last
place (ULP).
The sign bit and the zero case are excluded for brevity.

For a fixed input~$x$,
the quantized output $x_q(e, m; x)$ is differentiable
with respect to $e$ and $m$ when these quantities are
treated as real-valued parameters,
since the range bounds and bias vary continuously with
them.
The generalization of Eq.~\ref{eq:fp_quant} to continuous values and the derivation of the bit-widths is provided in Appendix~\ref{app:sec:float_conv},
and an example on how the quantization grid changes when using continuous values is shown in \HFigure{fig:quantization_grids}.

Although exponent and mantissa bit-widths could in
principle be learned independently,
this might produce configurations incompatible with
existing hardware.
To avoid this,
the format space is simplified by expressing each
supported format relative to a shared baseline.
The MXFP8, MXFP6, and MXFP4 representations correspond
to E4M3, E2M3, and E2M1,
respectively~\cite{ocpmx2023};
the E5M2 configuration for MXFP8 is not considered here,
as it is not typically used for the forward pass \cite{micikevicius2022fp8}.
We can define a generic MXFP($\beta$) configuration as
\begin{equation}
E(2{+}\beta)M(1{+}\beta),
\end{equation}
where $\beta$ is a learnable parameter.
When $\beta = 2$, we obtain the MXFP8 configuration (i.e., E4M3),
whilst $\beta=0$ gives us the baseline MXFP4 configuration.
During training,
$\beta$ may take any continuous value in $[0, 2]$,
permitting a smooth transition between configurations
and making the optimization less sensitive to abrupt
changes in the underlying representation.
This can be applied in the context of MXFP8/MXFP4
mixed precision configuration, where both $e$ and $m$
are optimized, and for the MXFP6/MXFP4 configuration,
where only $m$ is optimized.
Finally, tying weights and activations bit-widths reduces the problem
to a single learnable scalar per layer.

\subsection{Temperature-based
annealing}\label{subsec:temp_reg}

A continuous representation during training benefits
optimization stability,
but inference requires discrete bit-widths that map to
hardware-supported formats.
The central difficulty lies in managing this transition
without introducing instabilities;
at the same time, it is important to avoid discontinuities between
training-time and inference-time behavior.
A naive approach that applies a rounding operator with
an STE~\cite{bengio2013estimating} during each forward
pass can cause individual layers to oscillate between
MXFP4 and MXFP8 throughout training.
Such oscillations create abrupt changes in the loss
landscape and generally lead to worse final
performance,
as the experiments in
Section~\ref{subsec:results} confirm.

To avoid these oscillations,
we propose to smooth the bit-width offset through a
temperature-regulated sigmoid function.
The design goal is a mapping that behaves approximately
as an identity early in training,
allowing the optimizer to explore the continuous
bit-width space freely,
and that gradually sharpens into a step-like function
toward the end,
forcing convergence to one of the two discrete
endpoints, matching more closely the inference-time behaviour.
Our proposed formulation is:
\begin{equation}\label{eq:temp_sigmoid}
    \beta = F(\hat{\beta}, T)
    = \frac{2}{1 + e^{-T \cdot (\hat{\beta}/2 - 0.5)}},
\end{equation}
where $\beta$ is the offset parameter used to define the final bit-width as indicated in Sec. \ref{subsec:cont_param},
$\hat{\beta}$ is the raw learned offset, and $T$ is a
temperature coefficient that increases during
training.

When $T$ is small,
the sigmoid is nearly linear and the optimizer can
adjust bit-widths with minimal restriction.
As training proceeds,
$T$ increases and the function sharpens into a step
centered at $\beta = 1$.
Near the end of training,
offsets below~1 are pushed toward
$\beta = 0$ (MXFP4),
while offsets above~1 are pushed toward
$\beta = 2$
(MXFP8 or MXFP6, depending on the chosen format
range).
This annealing scheme ensures that the final precision
assignments are compatible with the discrete formats
supported by the target hardware,
while the gradual transition avoids the abrupt format
switches that characterize the rounding-based
alternative.

\subsection{Bit-width
regularization}\label{subsec:bit_reg}

Since higher precision generally reduces quantization error and
quantization error may limit how much the model task loss (e.g., cross-entropy loss) can be minimised,
a regularization term is required to balance
model quality against inference efficiency.
We consider two regularization strategies in this work:
a simple scaling penalty and a target-aware penalty.

\paragraph{Simple scaling penalty.}
The first strategy multiplies the current average
bit-width by a constant factor $\lambda$:
\begin{equation}\label{eq:bw_reg_simple}
    \mathcal{R_{s}} = \lambda
    \cdot \bar{b}_{\mathrm{current}}.
\end{equation}
This form has a straightforward interpretation,
with higher values of $\lambda$ leading to more aggressive quantization,
and vice-versa.
However, it provides only indirect
control over the final model configuration,
since it is not possible to predict apriori how a given
change in $\lambda$ will affect the resulting bit-width
allocation.

\paragraph{Target-aware penalty.}
The second strategy steers optimization toward a
specific target bit-width through the penalty
\begin{equation}\label{eq:bw_reg_target}
    \mathcal{R}_{t} = \max\!\left(0,\;
    \lambda \cdot \left(\bar{b}_{\mathrm{current}}
    - \bar{b}_{\mathrm{target}}\right)\right),
\end{equation}
where $\bar{b}_{\mathrm{current}}$ is the current
average bit-width and
$\bar{b}_{\mathrm{target}}$ is the desired target.
When the current average lies below the target,
the penalty vanishes and the bit-width is driven
solely by the task loss.
When the current average exceeds the target,
the penalty grows in proportion to the gap,
providing a direct incentive to reduce precision.
This formulation offers finer control over the
resulting configuration.
A practical limitation arises when
$\bar{b}_{\mathrm{current}}$ and
$\bar{b}_{\mathrm{target}}$ are close:
the penalty approaches zero and the gradient signal
weakens,
making precise convergence to the target dependent on
the magnitude of~$\lambda$.

Other regularization options could be considered,
based on the specific use-case (e.g., accounting
for stricter memory or compute constraints);
we leave this exploration for a future work,
while we present the results of the two strategies
presented above in Section~\ref{subsec:results}.

\subsubsection{Model-wise representative bit-width}
For both regularization losses,
we employ a model-wise representative bit-width for the entire model, $\bar{b}_{\mathrm{current}}$, for which we propose two formulations.

\paragraph{Simple average.} The first computes a simple unweighted average of the
per-layer bit-widths:
\begin{equation}\label{eq:avg_bw}
    \bar{b}_w = \frac{1}{N}\sum_{i=1}^{N} b_{w,i},
    \quad
    \bar{b}_a = \frac{1}{N}\sum_{i=1}^{N} b_{a,i},
    \quad
    \bar{b}_{\mathrm{current}}
    = \frac{\bar{b}_w + \bar{b}_a}{2},
\end{equation}
where $b_{w,i}$ and $b_{a,i}$ denote the weight and
activation bit-widths of the $i$-th layer,
and $N$ is the total number of quantized layers.

\paragraph{Weighted average.} The second formulation uses a weighted average in which
each layer's contribution is proportional to the number
of elements in its tensor.
Larger layers dominate the overall model footprint,
so weighting by size aligns the regularization
penalty with actual memory cost:
\begin{equation}\label{eq:weighted_avg_bw}
    \bar{b}_w^{\,\text{wt}}
    = \frac{\sum_{i=1}^{N} n_{w,i}
      \cdot b_{w,i}}
           {\sum_{i=1}^{N} n_{w,i}},
    \quad
    \bar{b}_a^{\,\text{wt}}
    = \frac{\sum_{i=1}^{N} n_{a,i}
      \cdot b_{a,i}}
           {\sum_{i=1}^{N} n_{a,i}},
    \quad
    \bar{b}_{\mathrm{current}}^{\,\text{wt}}
    = \frac{\bar{b}_w^{\,\text{wt}}
          + \bar{b}_a^{\,\text{wt}}}{2},
\end{equation}
where $n_{w,i}$ and $n_{a,i}$ denote the number of
elements in the weight and activation tensors of the
$i$-th layer, respectively.
The choice between these two formulations affects how
the precision budget is distributed:
the simple average treats all layers equally,
while the weighted average concentrates the penalty on
layers with the largest parameter counts.
We compare these approaches in Section~\ref{subsec:results}.

\section{Experimental Setting}\label{sec:exp_setting}

Experiments are conducted on three language models of
comparable scale:
Llama~3.2~1B~\cite{grattafiori2024llama3},
Qwen3~1.7B~\cite{yang2025qwen3},
and SmolLM2~1.7B~\cite{allal2025smollm2}.
We further validate that \OurAlgo scales to larger
models, up to 8~billion parameters,
in Appendix~\ref{app:sec:big_models}.
The training infrastructure we use is similar to the one proposed in SpinQuant~\cite{liu2024spinquant}, but we optimize rotations and bit-widths jointly.
The main difference, alongside all the modifications required to do learned bit-width assignment,
is that the calibration uses 3200 samples drawn from the FineWeb~\cite{penedo2024fineweb}
dataset and processed with a batch size of~8,
yielding 400 optimization steps.
This is a modest increase in training steps compared to the
SpinQuant training setup,
while still being considerably faster and more convenient
than full QAT training.

We evaluate our approach on perplexity computed on
WikiText-2~\cite{merity2017pointer},
and accuracy averaged over four zero-shot
reasoning benchmarks:
ARC-Challenge and ARC-Easy~\cite{clark2018arc},
HellaSwag~\cite{zellers2019hellaswag},
and WinoGrande~\cite{sakaguchi2020winogrande},
measured with the LightEval
framework~\cite{lighteval}.
We quantize all models using Brevitas \cite{brevitas}.

On the format side,
MXFP8/MXFP4 and MXFP6/MXFP4 are evaluated as the
two mixed-precision format configurations.

We also examine several design choices through
controlled ablations.
Tensor-size-weighted and simple-average formulations
of the bit-width regularization
(Section~\ref{subsec:bit_reg}) are compared,
as are the two regularization objectives
(simple scaling versus target-aware penalty).
Finally, dMX is compared against a greedy heuristic for layer selection,
in particular KL divergence-based pre-selected precision
assignments.
Finally, we compare \OurAlgo~with a discretized
variant that applies rounding with
an STE during each forward pass (ROUND+STE).
These results are presented in Appendix \ref{app:sec:ablation}.
For all the experiments described so far,
we use MXFP4 as starting configuration for all layers.

As baseline, we consider identical
quantization configuration
(i.e., MXFP4, MXFP6, or MXFP8)
uniformly across all layers.
All baselines use the same calibration setup as the
learned mixed-precision experiments;
the only difference is that their bit-widths remain
fixed throughout optimization.

In all figures,
each configuration is plotted against its average
bit-width computed from the per-layer bit-widths of
both weights and activations.
Unless stated otherwise,
average bit-width refers to the simple average across
layers.
Each point corresponds to a different target average
bit-width ranging from $4.1$ to $8$.

\paragraph{Hyper-parameter considerations.}
The proposed formulation introduces several
hyper-parameters.
The temperature schedule of
Section~\ref{subsec:temp_reg} requires initial and
final values for $T$,
as well as a ramp profile.
In all experiments,
the schedule follows an exponential ramp.
The initial and final values of $T$ are chosen to
approximate a near-linear mapping and a near-step
function, respectively.

A warm-up phase is also introduced:
for a fixed fraction of calibration steps,
denoted $T_{\mathrm{ratio}}$,
the temperature remains constant so that the mapping
stays approximately linear.
After this phase,
the exponential schedule begins and the bit-widths are
progressively discretized.
Experiments with different values of
$T_{\mathrm{ratio}}$ suggest that the method is
fairly robust to this choice.

Across several models,
effective settings were found with relatively modest
tuning effort.
More careful hyper-parameter search may yield further
improvements. We defer that exploration to future work.
The details about the hyper-parameters for our experiments
are detailed in Appendix \ref{app:sec:hyperparam}.

\subsection{Main Results}\label{subsec:results}
We first present the results associated to the MXFP8 and MXFP6 mixed precision configurations.
In the following tables we compare against the float baseline,
and the relevant MXFP configurations.

For the sake of brevity,
we only report results with target bit-width equal to $4.5, 5, 6$, and $8$ for MXFP8/MXFP4
mixed precision configuration, and $4.5, 5$, and $6$ for MXFP6/MXFP4 configurations.
We present the plot with all the different target bit-widths in the Appendix \ref{app:sec:ablation}.

Tables \ref{table:mxfp8_results}-\ref{table:mxfp6_results} show how even small increase in average bit-width
can lead to consistent improvements in terms of both perplexity and zero-shot accuracy.
Moreover, we are always able to closely match the desired target bit-width,
with the main exception being the highest bit-width target;
we believe this could be fixed with a more extensive hyper-parameter search
(e.g., adjusting the learning rate).

We further analyze which layers \OurAlgo most frequently keeps at high
precision, reporting the per-layer-type frequencies in
Appendix~\ref{app:sec:frequency}
(Tables~\ref{tab:freq_llama_mxfp8}--\ref{tab:freq_smollm2}).
Across models and configurations, the \texttt{down\_proj} and \texttt{v\_proj}
layers emerge as the most sensitive to quantization,
being consistently assigned the high-precision format.

\begin{table*}[ht]
\centering
\caption{We compare MXFP8/MXFP4 mixed precision versus three baselines: the base BF16 model, and quantization configurations where all layers are quantized to MXFP8 and MXFP4.
When using \OurAlgo, we report four different target bit-widths, using the average bit-width across all layers.}
\label{table:mxfp8_results}
\footnotesize
    \begin{tabular}{l l c c c}
    \toprule
    Model & Type & \begin{tabular}{c} Average Bit-Width \end{tabular} & PPL & Zero-Shot \\
    \midrule
    \multirow{7}{*}{Llama 3.2 1B} & Float & \textit{BF16} & 8.94 & 51.53 \\
     & MXFP4 & 4.0 & 11.68 & 47.82 \\
     & MXFP8 & 8.0 & 9.15 & 51.39 \\
    \cmidrule{2-5}
     & \multirow{4}{*}{dMX} & 4.57 & 11.02 & 48.11 \\
     &  & 5.11 & 10.60 & 49.27 \\
     &             & 6.04 & 9.83 & 49.61 \\
     &             & 8.0 & 9.19 & 51.46 \\
    \midrule
    \multirow{7}{*}{SmolLM2 1.7B} & Float & \textit{BF16} & 7.61 & 59.52 \\
     & MXFP4 & 4.0 & 10.28 & 53.42 \\
     & MXFP8 & 8.0 & 7.86 & 58.69 \\
    \cline{2-5}
     & \multirow{4}{*}{dMX} & 4.53 & 9.33 & 54.41 \\
     &  & 5.09 & 8.92 & 55.36 \\
     &             & 6.06 & 8.26 & 57.28 \\
     &             & 7.29 & 7.96 & 57.90 \\
    \midrule
    \multirow{7}{*}{Qwen3 1.7B} & Float & \textit{BF16} & 15.74 & 54.12 \\
     & MXFP4 & 4.0 & 12.57 & 51.15 \\
     & MXFP8 & 8.0 & 10.50 & 54.08 \\
    \cline{2-5}
     & \multirow{4}{*}{dMX} & 4.53 & 11.91 & 52.20 \\
     &  & 5.02 & 11.58 & 51.62\\
     &             & 5.90 & 11.02 & 53.74 \\
     &             & 7.65 & 10.47 & 54.09 \\
    \bottomrule
    \end{tabular}
\end{table*}

\begin{table}[ht]
\centering
\caption{We compare MXFP6/MXFP4 mixed precision versus three baselines: the base BF16 model, and quantization configurations where all layers are quantized to MXFP6 and MXFP4.
When using \OurAlgo, we report three different target bit-widths, using the average bit-width across all layers.}
\label{table:mxfp6_results}
\footnotesize
\begin{tabular}{l l c c c}
\toprule
Model & Type & \begin{tabular}{c} Average Bit-Width \end{tabular} & PPL & Zero-Shot \\
\midrule
\multirow{7}{*}{Llama 3.2 1B} & Float & \textit{BF16} & 8.94 & 51.53 \\
 & MXFP4 & 4.0 & 11.68 & 47.82 \\
 & MXFP6 & 6.0 & 9.16 & 51.39 \\
\cmidrule{2-5}
 & \multirow{3}{*}{dMX} & 4.52 & 10.76 & 49.43 \\
 &  & 5.02 & 10.02 & 50.26 \\
 &             & 5.98 & 9.48 & 50.86 \\
\midrule
\multirow{7}{*}{SmolLM2 1.7B} & Float & \textit{BF16} & 7.61 & 59.52 \\
 & MXFP4 & 4.0 & 10.28 & 53.42 \\
 & MXFP6 & 6.0 & 7.81 & 58.18 \\
\cline{2-5}
 & \multirow{3}{*}{dMX} & 4.52 & 9.78 & 56.40 \\
 &  & 5.06 & 8.50 & 57.52 \\
 &             & 5.54 & 8.26 & 58.25 \\
\midrule
\multirow{7}{*}{Qwen3 1.7B} & Float & \textit{BF16} & 15.74 & 54.12 \\
 & MXFP4 & 4.0 & 12.57 & 51.15 \\
 & MXFP6 & 6.0 & 10.67 & 54.05 \\
\cmidrule{2-5}
 & \multirow{3}{*}{dMX} & 4.52 & 11.53 & 52.90 \\
 &  & 5.03 & 11.01 & 54.14 \\
 &             & 5.69 & 10.64 & 54.86 \\
\bottomrule
\end{tabular}
\end{table}

\subsection{Ablation Studies}\label{subsec:ablation}
For the ablation studies,
we present the plots with all the different target bit-widths to have a better
understanding of the various comparisons.

\paragraph{Regularization function comparison.}

\begin{figure*}[h!]
    \input{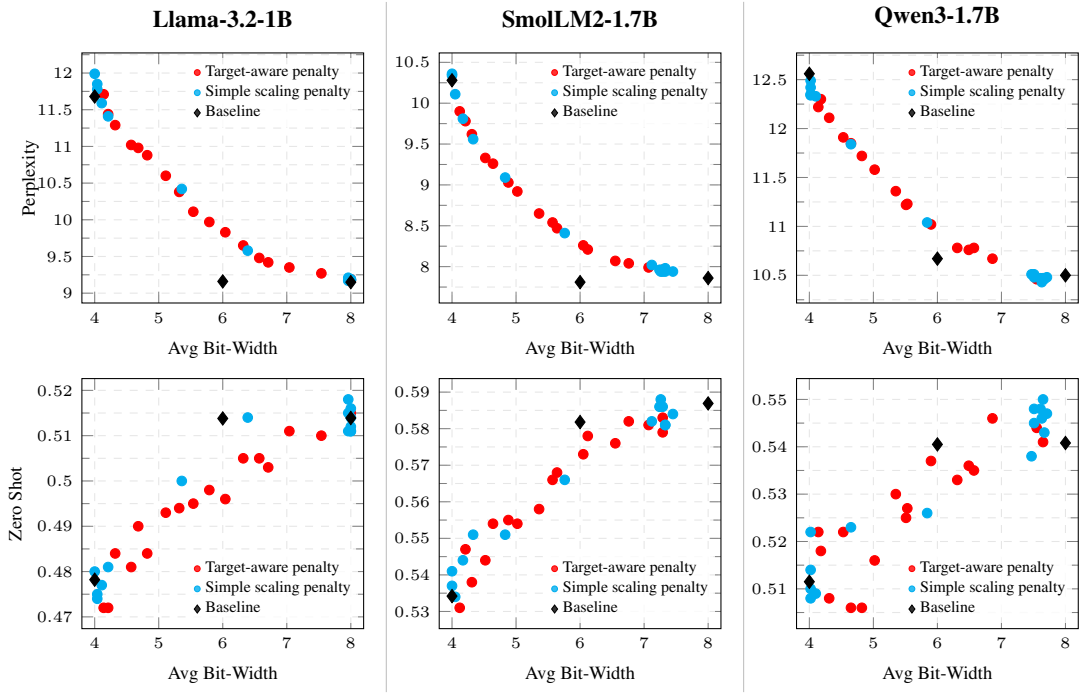}
    \pgfplotstableread[col sep=comma]{\csvAone}\dataAone
\pgfplotstableread[col sep=comma]{\csvBone}\dataBone
\pgfplotstableread[col sep=comma]{\csvCone}\dataCone

\ifshowSeriesB
  \pgfplotstableread[col sep=comma]{\csvAtwo}\dataAtwo
  \pgfplotstableread[col sep=comma]{\csvBtwo}\dataBtwo
  \pgfplotstableread[col sep=comma]{\csvCtwo}\dataCtwo
\fi

\pgfplotstableread[col sep=comma]{\refCsvA}\refDataA
\pgfplotstableread[col sep=comma]{\refCsvB}\refDataB
\pgfplotstableread[col sep=comma]{\refCsvC}\refDataC

\begin{tikzpicture}
  \begin{groupplot}[
      group style={
        group size=3 by 2,
        horizontal sep=1.0cm,
        vertical sep=1.0cm,
      },
      width=0.38\textwidth, height=0.35\textwidth,
      xmin=3.8, xmax=8.2,
      grid=both,
      grid style={dashed, gray!20},
      every axis plot/.append style={only marks, mark=*, mark size=1.8pt},
      legend cell align=left,
      legend style={
        font=\tiny,
        row sep=-2pt,
        draw=none,
        fill=white,
        fill opacity=0.8,
        text opacity=1,
        inner sep=0.5pt,
        legend image post style={scale=0.6},
      },
      title style={font=\small\bfseries},
      label style={font=\scriptsize},
      tick label style={font=\tiny},
      ylabel style={at={(-0.12,0.5)}},
      ytick distance=0.5,          
      minor y tick num=1,
    ]


    \nextgroupplot[
      title={\modelTitleA},
      xlabel={\xaxislabel},
      ylabel={Perplexity},
      legend style={at={(0.98,0.98)}, anchor=north east},
    ]
    \addplot[colorA] table[x=bitwidth, y=ppl] {\dataAone};
      \addlegendentry{\legendA}
    \ifshowSeriesB
      \addplot[colorB] table[x=bitwidth, y=ppl] {\dataAtwo};
        \addlegendentry{\legendB}
    \fi
    \addplot[black, mark=diamond*, mark size=2.5pt]
      table[x=bitwidth, y=ppl] {\refDataA};
      \addlegendentry{\legendRef}

    \nextgroupplot[
      title={\modelTitleB},
      xlabel={\xaxislabel},
      legend style={at={(0.98,0.98)}, anchor=north east},
    ]
    \addplot[colorA] table[x=bitwidth, y=ppl] {\dataBone};
      \addlegendentry{\legendA}
    \ifshowSeriesB
      \addplot[colorB] table[x=bitwidth, y=ppl] {\dataBtwo};
        \addlegendentry{\legendB}
    \fi
    \addplot[black, mark=diamond*, mark size=2.5pt]
      table[x=bitwidth, y=ppl] {\refDataB};
      \addlegendentry{\legendRef}

    \nextgroupplot[
      title={\modelTitleC},
      xlabel={\xaxislabel},
      legend style={at={(0.98,0.98)}, anchor=north east},
    ]
    \addplot[colorA] table[x=bitwidth, y=ppl] {\dataCone};
      \addlegendentry{\legendA}
    \ifshowSeriesB
      \addplot[colorB] table[x=bitwidth, y=ppl] {\dataCtwo};
        \addlegendentry{\legendB}
    \fi
    \addplot[black, mark=diamond*, mark size=2.5pt]
      table[x=bitwidth, y=ppl] {\refDataC};
      \addlegendentry{\legendRef}


    \nextgroupplot[
      xlabel={\xaxislabel},
      ylabel={Zero Shot},
      ylabel style={at={(-0.18,0.5)}},
      ytick distance=0.01,
      minor y tick num=1,
      legend style={at={(0.98,0.02)}, anchor=south east},
    ]
    \addplot[colorA] table[x=bitwidth, y=zs] {\dataAone};
      \addlegendentry{\legendA}
    \ifshowSeriesB
      \addplot[colorB] table[x=bitwidth, y=zs] {\dataAtwo};
        \addlegendentry{\legendB}
    \fi
    \addplot[black, mark=diamond*, mark size=2.5pt]
      table[x=bitwidth, y=zs] {\refDataA};
      \addlegendentry{\legendRef}

    \nextgroupplot[
      xlabel={\xaxislabel},
      ytick distance=0.01,
      minor y tick num=1,
      legend style={at={(0.98,0.02)}, anchor=south east},
    ]
    \addplot[colorA] table[x=bitwidth, y=zs] {\dataBone};
      \addlegendentry{\legendA}
    \ifshowSeriesB
      \addplot[colorB] table[x=bitwidth, y=zs] {\dataBtwo};
        \addlegendentry{\legendB}
    \fi
    \addplot[black, mark=diamond*, mark size=2.5pt]
      table[x=bitwidth, y=zs] {\refDataB};
      \addlegendentry{\legendRef}

    \nextgroupplot[
      xlabel={\xaxislabel},
      ytick distance=0.01,
      minor y tick num=1,
      legend style={at={(0.98,0.02)}, anchor=south east},
    ]
    \addplot[colorA] table[x=bitwidth, y=zs] {\dataCone};
      \addlegendentry{\legendA}
    \ifshowSeriesB
      \addplot[colorB] table[x=bitwidth, y=zs] {\dataCtwo};
        \addlegendentry{\legendB}
    \fi
    \addplot[black, mark=diamond*, mark size=2.5pt]
      table[x=bitwidth, y=zs] {\refDataC};
      \addlegendentry{\legendRef}

  \end{groupplot}

  \coordinate (sep12top) at ($ (group c1r1.east)!0.3!(group c2r1.west) $);
  \coordinate (sep12bot) at ($ (group c1r2.east)!0.3!(group c2r2.west) $);
  \coordinate (sep23top) at ($ (group c2r1.east)!0.3!(group c3r1.west) $);
  \coordinate (sep23bot) at ($ (group c2r2.east)!0.3!(group c3r2.west) $);
  \draw[gray!50, thin]
    (sep12top |- group c1r1.outer north) -- (sep12bot |- group c1r2.outer south);
  \draw[gray!50, thin]
    (sep23top |- group c2r1.outer north) -- (sep23bot |- group c2r2.outer south);
\end{tikzpicture}
\caption{Comparison of the target-aware penalty and
    the simple scaling penalty for MXFP8--MXFP4
    mixed-precision quantization.
    Red: target-aware penalty
    ($\mathcal{R}_t$).
    Cyan: simple scaling penalty
    ($\mathcal{R}_s$).
    Black diamonds: homogeneous MXFP baselines.
    Top row: perplexity (lower is better).
    Bottom row: average zero-shot accuracy
    (higher is better).
    The target-aware penalty produces a smoother
    spread of operating points across the bit-width
    range.}
\label{fig:simple_loss}
\end{figure*}

\HFigure{fig:simple_loss} compares the two
regularization objectives described in
Section~\ref{subsec:bit_reg}.
The target-aware penalty ($\mathcal{R}_t$) produces
a denser and more evenly distributed set of operating
points across the bit-width range,
particularly for Llama~3.2~1B and SmolLM2~1.7B.
The simple scaling penalty ($\mathcal{R}_s$) tends to
cluster configurations at the extremes of the
bit-width range.

As expected the target-aware
formulation offers more predictable control over the
final bit-width allocation,
and the extra constraints on the target bit-width does not
to negatively impact the training.

For the rest of the experiments,
we use the target-aware cost function,
since it allows a better understanding
of the transition between bit-width configurations.

\paragraph{Cost function comparison.}
As described in Section~\ref{subsec:bit_reg},
we investigate two cost function strategies, a simple
average or a tensor-size-weighted average of the
per-layer bit-widths.
Under the weighted formulation,
larger layers receive a stronger penalty,
which is expected to push the optimizer toward assigning
them lower precision.

\begin{figure*}
    \centering
    \input{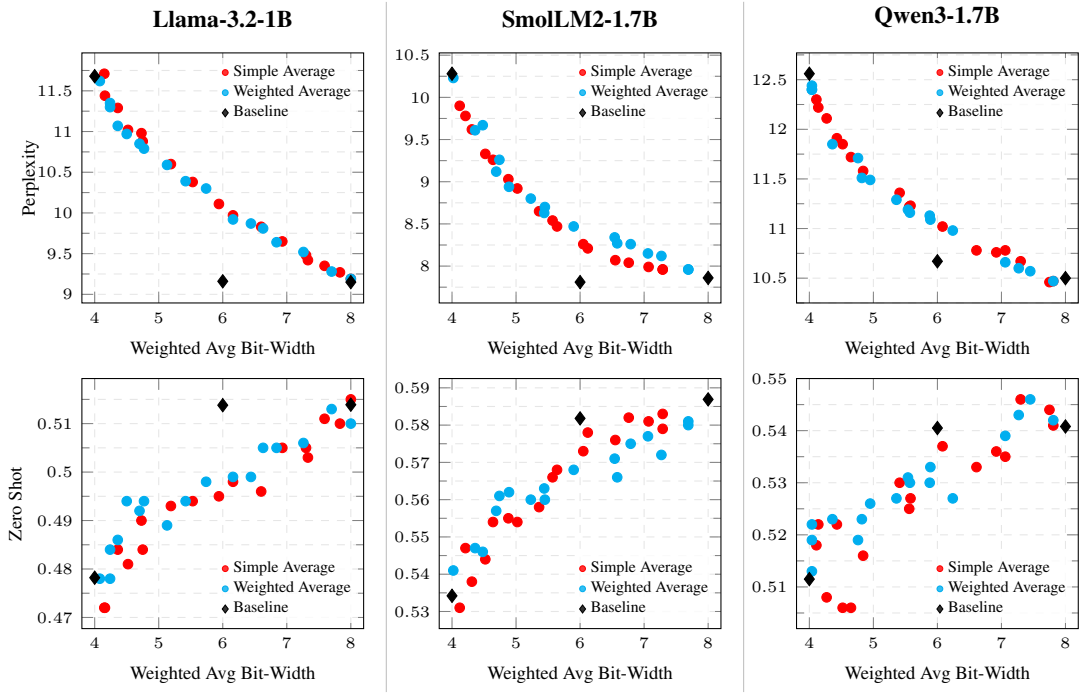}
    \pgfplotstableread[col sep=comma]{\csvAone}\dataAone
\pgfplotstableread[col sep=comma]{\csvBone}\dataBone
\pgfplotstableread[col sep=comma]{\csvCone}\dataCone

\ifshowSeriesB
  \pgfplotstableread[col sep=comma]{\csvAtwo}\dataAtwo
  \pgfplotstableread[col sep=comma]{\csvBtwo}\dataBtwo
  \pgfplotstableread[col sep=comma]{\csvCtwo}\dataCtwo
\fi

\pgfplotstableread[col sep=comma]{\refCsvA}\refDataA
\pgfplotstableread[col sep=comma]{\refCsvB}\refDataB
\pgfplotstableread[col sep=comma]{\refCsvC}\refDataC

\begin{tikzpicture}
  \begin{groupplot}[
      group style={
        group size=3 by 2,
        horizontal sep=1.0cm,
        vertical sep=1.0cm,
      },
      width=0.38\textwidth, height=0.35\textwidth,
      xmin=3.8, xmax=8.2,
      grid=both,
      grid style={dashed, gray!20},
      every axis plot/.append style={only marks, mark=*, mark size=1.8pt},
      legend cell align=left,
      legend style={
        font=\tiny,
        row sep=-2pt,
        draw=none,
        fill=white,
        fill opacity=0.8,
        text opacity=1,
        inner sep=0.5pt,
        legend image post style={scale=0.6},
      },
      title style={font=\small\bfseries},
      label style={font=\scriptsize},
      tick label style={font=\tiny},
      ylabel style={at={(-0.12,0.5)}},
      ytick distance=0.5,          
      minor y tick num=1,
    ]


    \nextgroupplot[
      title={\modelTitleA},
      xlabel={\xaxislabel},
      ylabel={Perplexity},
      legend style={at={(0.98,0.98)}, anchor=north east},
    ]
    \addplot[colorA] table[x=bitwidth, y=ppl] {\dataAone};
      \addlegendentry{\legendA}
    \ifshowSeriesB
      \addplot[colorB] table[x=bitwidth, y=ppl] {\dataAtwo};
        \addlegendentry{\legendB}
    \fi
    \addplot[black, mark=diamond*, mark size=2.5pt]
      table[x=bitwidth, y=ppl] {\refDataA};
      \addlegendentry{\legendRef}

    \nextgroupplot[
      title={\modelTitleB},
      xlabel={\xaxislabel},
      legend style={at={(0.98,0.98)}, anchor=north east},
    ]
    \addplot[colorA] table[x=bitwidth, y=ppl] {\dataBone};
      \addlegendentry{\legendA}
    \ifshowSeriesB
      \addplot[colorB] table[x=bitwidth, y=ppl] {\dataBtwo};
        \addlegendentry{\legendB}
    \fi
    \addplot[black, mark=diamond*, mark size=2.5pt]
      table[x=bitwidth, y=ppl] {\refDataB};
      \addlegendentry{\legendRef}

    \nextgroupplot[
      title={\modelTitleC},
      xlabel={\xaxislabel},
      legend style={at={(0.98,0.98)}, anchor=north east},
    ]
    \addplot[colorA] table[x=bitwidth, y=ppl] {\dataCone};
      \addlegendentry{\legendA}
    \ifshowSeriesB
      \addplot[colorB] table[x=bitwidth, y=ppl] {\dataCtwo};
        \addlegendentry{\legendB}
    \fi
    \addplot[black, mark=diamond*, mark size=2.5pt]
      table[x=bitwidth, y=ppl] {\refDataC};
      \addlegendentry{\legendRef}


    \nextgroupplot[
      xlabel={\xaxislabel},
      ylabel={Zero Shot},
      ylabel style={at={(-0.18,0.5)}},
      ytick distance=0.01,
      minor y tick num=1,
      legend style={at={(0.98,0.02)}, anchor=south east},
    ]
    \addplot[colorA] table[x=bitwidth, y=zs] {\dataAone};
      \addlegendentry{\legendA}
    \ifshowSeriesB
      \addplot[colorB] table[x=bitwidth, y=zs] {\dataAtwo};
        \addlegendentry{\legendB}
    \fi
    \addplot[black, mark=diamond*, mark size=2.5pt]
      table[x=bitwidth, y=zs] {\refDataA};
      \addlegendentry{\legendRef}

    \nextgroupplot[
      xlabel={\xaxislabel},
      ytick distance=0.01,
      minor y tick num=1,
      legend style={at={(0.98,0.02)}, anchor=south east},
    ]
    \addplot[colorA] table[x=bitwidth, y=zs] {\dataBone};
      \addlegendentry{\legendA}
    \ifshowSeriesB
      \addplot[colorB] table[x=bitwidth, y=zs] {\dataBtwo};
        \addlegendentry{\legendB}
    \fi
    \addplot[black, mark=diamond*, mark size=2.5pt]
      table[x=bitwidth, y=zs] {\refDataB};
      \addlegendentry{\legendRef}

    \nextgroupplot[
      xlabel={\xaxislabel},
      ytick distance=0.01,
      minor y tick num=1,
      legend style={at={(0.98,0.02)}, anchor=south east},
    ]
    \addplot[colorA] table[x=bitwidth, y=zs] {\dataCone};
      \addlegendentry{\legendA}
    \ifshowSeriesB
      \addplot[colorB] table[x=bitwidth, y=zs] {\dataCtwo};
        \addlegendentry{\legendB}
    \fi
    \addplot[black, mark=diamond*, mark size=2.5pt]
      table[x=bitwidth, y=zs] {\refDataC};
      \addlegendentry{\legendRef}

  \end{groupplot}

  \coordinate (sep12top) at ($ (group c1r1.east)!0.3!(group c2r1.west) $);
  \coordinate (sep12bot) at ($ (group c1r2.east)!0.3!(group c2r2.west) $);
  \coordinate (sep23top) at ($ (group c2r1.east)!0.3!(group c3r1.west) $);
  \coordinate (sep23bot) at ($ (group c2r2.east)!0.3!(group c3r2.west) $);
  \draw[gray!50, thin]
    (sep12top |- group c1r1.outer north) -- (sep12bot |- group c1r2.outer south);
  \draw[gray!50, thin]
    (sep23top |- group c2r1.outer north) -- (sep23bot |- group c2r2.outer south);
\end{tikzpicture}
    \caption{Comparison of simple-average and
    tensor-size-weighted bit-width regularization
    for MXFP8/MXFP4 mixed-precision quantization.
    Red: simple average.
    Cyan: weighted average.
    Black diamonds: homogeneous MXFP baselines.
    Top row: perplexity
    (lower is better).
    Bottom row: average zero-shot accuracy
    (higher is better).
    The simple-average formulation tends to reach
    lower perplexity and spans a wider range of
    operating points.}
    \label{fig:average_vs_weighted_weighted}
\end{figure*}

\HFigure{fig:average_vs_weighted_weighted} compares
the two strategies.
In all panels the $x$-axis reports the same weighted
average bit-width,
defined as the average bit-width across layers,
weighted by the weight tensor size in each layer.
The results for the two loss functions results largely overlap,
showing how the user can easily adapt the method
to match their use cases.

\begin{figure*}
    \centering
    \input{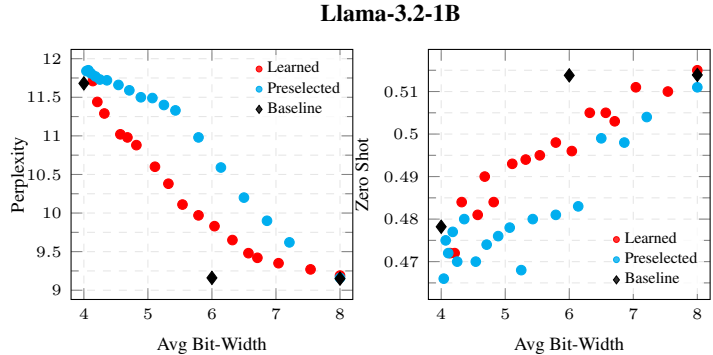}
    \pgfplotstableread[col sep=comma]{\csvAone}\dataAone

\ifshowSeriesB
  \pgfplotstableread[col sep=comma]{\csvAtwo}\dataAtwo

\fi

\pgfplotstableread[col sep=comma]{\refCsvA}\refDataA

\begin{tikzpicture}
  \begin{groupplot}[
      group style={
        group size=2 by 1,
        horizontal sep=1.0cm,
        vertical sep=1.0cm,
      },
      width=0.38\textwidth, height=0.35\textwidth,
      xmin=3.8, xmax=8.2,
      grid=both,
      grid style={dashed, gray!20},
      every axis plot/.append style={only marks, mark=*, mark size=1.8pt},
      legend cell align=left,
      legend style={
        font=\tiny,
        row sep=-2pt,
        draw=none,
        fill=white,
        fill opacity=0.8,
        text opacity=1,
        inner sep=0.5pt,
        legend image post style={scale=0.6},
      },
      title style={font=\small\bfseries},
      label style={font=\scriptsize},
      tick label style={font=\tiny},
      ylabel style={at={(-0.12,0.5)}},
      ytick distance=0.5,          
      minor y tick num=1,
    ]


    \nextgroupplot[
      xlabel={\xaxislabel},
      ylabel={Perplexity},
      legend style={at={(0.98,0.98)}, anchor=north east},
    ]
    \addplot[colorA] table[x=bitwidth, y=ppl] {\dataAone};
      \addlegendentry{\legendA}
    \ifshowSeriesB
      \addplot[colorB] table[x=bitwidth, y=ppl] {\dataAtwo};
        \addlegendentry{\legendB}
    \fi
    \addplot[black, mark=diamond*, mark size=2.5pt]
      table[x=bitwidth, y=ppl] {\refDataA};
      \addlegendentry{\legendRef}

    \nextgroupplot[
      xlabel={\xaxislabel},
      ylabel={Zero Shot},
      ylabel style={at={(-0.18,0.5)}},
      ytick distance=0.01,
      minor y tick num=1,
      legend style={at={(0.98,0.02)}, anchor=south east},
    ]
    \addplot[colorA] table[x=bitwidth, y=zs] {\dataAone};
      \addlegendentry{\legendA}
    \ifshowSeriesB
      \addplot[colorB] table[x=bitwidth, y=zs] {\dataAtwo};
        \addlegendentry{\legendB}
    \fi
    \addplot[black, mark=diamond*, mark size=2.5pt]
      table[x=bitwidth, y=zs] {\refDataA};
      \addlegendentry{\legendRef}

  \end{groupplot}
    \node[above=7pt, font=\small\bfseries] at ($ (group c1r1.north)!0.5!(group c2r1.north) $)
    {\modelTitleA};
\end{tikzpicture}
\caption{Learned bit-width optimization vs.\
    KL divergence-based pre-selected layer precision
    for Llama~3.2~1B.
    Red: learned.
    Cyan: pre-selected.
    Black diamonds: homogeneous MXFP baselines.
    The $x$-axis reports the weighted average
    bit-width.
    Left: perplexity
    (lower is better).
    Right: average zero-shot accuracy
    (higher is better).
    The learned approach achieves lower perplexity
    and higher accuracy at intermediate bit-widths.}
\label{fig:llama1_preselected_horizontal}
\end{figure*}

\paragraph{Learned vs.\ pre-selected layer precision.}
An alternative to gradient-based bit-width learning is
sensitivity-based selection.
Each layer is quantized to lower precision in
isolation,
and the KL divergence between the output distributions
of the full-precision model and the
single-layer-quantized model is measured.
Layers with larger KL divergence are treated as more
sensitive and assigned higher precision first.
\HFigure{fig:llama1_preselected_horizontal}
presents the comparison for Llama~3.2~1B.

At the lowest average bit-widths,
the learned and pre-selected approaches perform
similarly because nearly all layers are forced to
MXFP4.
The few layers that are quantized to MXFP8
have a high sensitivity to quantization,
which can also be identified through KL divergence.

At intermediate bit-widths,
roughly between 5.0 and 7.0,
the sensitivity approximation becomes less reliable,
causing
the pre-selected curve to remain relatively flat in this
region, while
the learned assignment achieves noticeably lower
perplexity.
The same trend appears in zero-shot accuracy,
where the learned approach outperforms the pre-selected
baseline across most of the intermediate range.
At high bit-widths both methods converge toward the
MXFP8 baseline.
This advantage at intermediate operating points is
consistent with the expectation that end-to-end
optimization captures cross-layer dependencies that
isolated sensitivity estimates miss.

\section{Related Work}\label{sec:related}

\paragraph{Layer-wise bit-width selection.}
Determining an optimal bit-width allocation across
layers is a combinatorial optimization problem whose
search space grows exponentially with the number of
layers and candidate
precisions~\cite{wu2018mixed,cohen2025gradient_free}.
Early sensitivity-based approaches estimate per-layer
importance through proxies such as the layer's Hessian spectrum in HAWQ~\cite{dong2019hawq} or
reinforcement-learning-based search over
hardware-oriented objectives in
HAQ~\cite{wang2019haq}.
Subsequent refinements have replaced or extended these
proxies:
HAWQ-V3~\cite{yao2021hawqv3} casts the allocation as
an integer linear program,
and Chen et al.~\cite{chen2021constrained} formulate it
as a constrained knapsack problem.
In the context of LLMs,
APTQ~\cite{guan2024aptq} combines Hessian-trace
sensitivity with attention-aware metrics,
ResQ~\cite{saxena2024resq} separates principal
components through PCA to isolate outlier structure.
A common finding across these works is that
mixed-precision quantization is often essential for
pushing below four bits per parameter without
unacceptable degradation,
yet they focus almost entirely on integer formats.

\paragraph{Gradient-based bit-width learning.}
Learning bit-widths through gradient descent has been previously explored for integer
quantization~\cite{wu2018mixed,cai2020edmips}.
FracBits~\cite{yang2021fracbits} uses linear interpolation between adjacent integer bit-widths, while requiring a two-stage process for effective mixed-precision deployment.
while SDQ~\cite{huang2022sdq} employs auxiliary learned parameters that represents the probability of selecting a specific bit-width, with no continuous transition between values.
BSQ~\cite{yang2021bsq} recasts mixed-precision
assignment as bit-level sparsity,
and Q-ViT~\cite{li2022qvit} extends differentiable
precision search to vision transformers.
On the discrete-selection side,
HMQ~\cite{habi2020hmq} relies on a Gumbel-Softmax
relaxation~\cite{jang2016categorical,
maddison2017concrete},
Bayesian Bits~\cite{vanbaalen2020bayesianbits} unifies
quantization and pruning through stochastic gates,
and Uhlich et al.~\cite{uhlich2020mixed} demonstrate
that the parameterization of the quantizer itself
significantly affects optimization stability.
All of these methods target integer formats.
Floating-point and MX-based formats are becoming
the more prevalent for LLM deployment on recent
hardware~\cite{rouhani2023microscaling, micikevicius2022fp8},
which motivates extending gradient-based techniques
to this setting.

\paragraph{Floating-point quantization.}

Sensitivity-based approaches could in principle
be extended to floating-point formats,
For example, MicroMix~\cite{liu2025micromix} performs channel-wise assignment across MX formats,
using an heuristic approach to perform precision assignment.
Gradient-based methods are harder to transfer directly
because of the floating-point representation.
To the best of current knowledge,
no prior work has addressed this problem through a
gradient-based method that learns floating-point format
selection while using a continuous parameterization to
interpolate smoothly between hardware-supported formats
during calibration.

\section{Conclusion and Future Works}\label{sec:conclusion}

This work presented \OurAlgo,
a differentiable framework for learning per-layer
floating-point bit-widths, focusing on the MX format family.
The method parameterizes each layer's precision through
a single continuous offset that interpolates between
different hardware-supported MXFP formats,
and uses a temperature-based annealing schedule to
progressively discretize the learned offsets.

Our experiments on Llama~3.2~1B, Qwen3~1.7B, and
SmolLM2~1.7B demonstrated that learned mixed-precision
assignments consistently outperform both homogeneous
MXFP baselines and KL divergence-based heuristics,
particularly at intermediate bit-widths where the
precision budget must be allocated carefully across
layers.
We further showed that these benefits carry over to
larger models, up to 8~billion parameters,
as detailed in Appendix~\ref{app:sec:big_models}.

Ablation studies examined the roles of
initialization strategy,
format ranges,
and regularization formulations,
providing guidance for practical deployment.

Several directions remain open for future work.
Extending the approach to
mixture-of-experts models,
where layer sensitivities can vary more
dramatically~\cite{huang2025mixture},
is a natural next step.
A more systematic hyper-parameter study,
including alternative temperature schedules and
regularization forms,
may also improve robustness across a wider range of
models and deployment targets.

Although we have investigated this approach
in the context of gradient-based PTQ,
\OurAlgo~lends itself to applications in other domains,
such as parameter-efficient fine tuning (PEFT),
or QAT setups.

\section*{Acknowledgments}

We are grateful to Michaela Blott and Ralph Wittig for their continued support
throughout this project. We also thank Mukund Athmaram, who proposed ideas for
the effective presentation of our results and carefully proofread the
manuscript. Their contributions substantially improved this work.

\bibliographystyle{unsrt}
\bibliography{references}
\newpage
\appendix
\section{Continuous and Differentiable Float Conversion Function}
\label{app:sec:float_conv}

This appendix specifies how $\mathit{ss}$ in Eq.~\ref{eq:fp_quant} is computed
from each element's magnitude, mantissa bit-width
$m$, and exponent bias $b$.
Letting $\tilde{x} = x/s$ denote the magnitude-scaled element,
Eqs.~\ref{eq:fp_range} and ~\ref{eq:fp_quant} are reproduced here for reference:
\begin{equation}\label{eq:fp_range_rep}
    q_{\max}^{(\mathrm{FP})} = (2 - 2^{-m})
    \cdot 2^{\,2^{e} - b - 1},
    \qquad
    q_{\min}^{(\mathrm{FP})}
    = -q_{\max}^{(\mathrm{FP})}.
\end{equation}

\begin{equation}\label{eq:fp_quant_rep}
    x_q
    = \operatorname{clip}\!\left(
        \mathit{ss}(\tilde{x}) \left\lfloor
            \frac{\tilde{x}}{\mathit{ss}(\tilde{x})}
        \right\rceil;\;
        q_{\min}^{(\mathrm{FP})},\,
        q_{\max}^{(\mathrm{FP})}
    \right) 
\end{equation}

where $s$ is a coarse scale.

The purpose of $\mathit{ss}$ is to represent the mantissa part of the floating-point value
as an integer, permitting quantization via a standard rounding operator.
The resulting expression is differentiable in $m$ and $e$
everywhere except in the case of rounding operations.
These are handled by straight-through estimator,
formally defined in Section~\ref{app:subsec:gradients}.

\paragraph{Inputs and assumptions.}
Recall $\tilde{x}$ from above, with mantissa bit-width
$m$, exponent bit-width $e$, and exponent bias $b = 2^{e-1} - 1$.

The subnormal exponent floor is
\begin{equation}
\label{eq:appfp:emin}
e_{\min} \;=\; 1 - b - m,
\end{equation}
the exponent of the unit in the last place (ULP) is defined as
\begin{equation}
\label{eq:appfp:exponent}
\eta(\tilde{x})
\;=\;
\max\!\Bigl(\bigl\lfloor \log_2(|\tilde{x}|)
            \bigr\rfloor - m,\; e_{\min}\Bigr),
\end{equation}
and the per-element power-of-two scale is
\begin{equation}
\label{eq:appfp:ss}
\mathit{ss}(\tilde{x}) \;=\; 2^{\,\eta(\tilde{x})}.
\end{equation}

The sign bit and the zero case ($x = 0$) are excluded for brevity;

\paragraph{Interpretation.}
For nonzero $\tilde{x}$, the quantity
$\lfloor \log_2 \tilde{x} \rfloor$ is the binary
exponent of $\tilde{x}$, so $\lfloor \log_2 \tilde{x}
\rfloor - m$ is the exponent of the ULP of a normalized floating-point number with
$m$ mantissa bits at that magnitude.
Clamping the exponent below by $e_{\min}$ implements
the IEEE-754-style \cite{IEEE754} \emph{subnormal} regime: once the
true exponent falls below $1 - b$, the ULP stays fixed
at $2^{e_{\min}}$ and mantissa precision degrades toward zero.
The mantissa bit-width $m$ enters $\mathit{ss}$ through
both $\eta(\tilde{x})$ and $e_{\min}$, while the exponent
bias $b$ enters through $e_{\min}$; these are the entry
points for the gradient analysis in Section~\ref{app:subsec:gradients}.

\paragraph{From discrete to continuous.}
At deployment, hardware imposes the constraint $m \in \mathbb{Z}^{+}$ and
$e \in \mathbb{Z}^{+}$. However, Eqs.~\ref{eq:fp_range_rep} and~\ref{eq:fp_quant_rep}
remain well defined for $m, e \in \mathbb{R}^{+}$, and no mathematical constraint
forces integer values during training. Accordingly, $m$ and $e$ are treated as
real-valued parameters throughout calibration, subject to the range constraints
described in Section~\ref{subsec:cont_param}.

\paragraph{Continuous floating-point interpretation.}
The output of the continuous floating-point conversion function exhibits the
following properties:
\begin{itemize}
    \item It is a non-uniform \emph{staircase} function whose distinct quantization
          levels need not be powers of two.
    \item The number of distinct staircase step sizes, determined by changes in the
          exponent, likewise need not align with powers of two.
\end{itemize}
As $m$ increases, the number of quantization levels within each step-size increases;
as $e$ increases, the number of distinct step sizes increases.
See \HFigure{fig:quantization_grids} for an example grid and
panel~\ref{fig:appfp:bitwidth_gradients:a} of
\HFigure{fig:appfp:bitwidth_gradients} for the forward staircase
at $e = 2$, $m = 2$.

\subsection{Gradients via the straight-through estimator}
\label{app:subsec:gradients}

We now derive the gradients of the unscaled inner quantizer $Q(\tilde{x})=x_q$
with respect to the mantissa bit-width $m$, the exponent
bit-width $e$, and the shared offset $\beta$ that parameterizes
both via $m = 1+\beta$, $e = 2+\beta$
(Section~\ref{subsec:cont_param}).
For this analysis, we consider $s$ to be constant in $m$ and
$e$, so the bit-width gradients of $x_q$ and of $Q$ differ only
by the outer $\operatorname{sign}(x)\cdot s$ factor, and we
focus on $Q$ throughout.

\paragraph{STE convention.}
We adopt the straight-through estimator of
\cite{bengio2013estimating}, following the convention used in
TQT~\cite{jain2020trained}:
\begin{equation}
\label{eq:appfp:ste}
\frac{\partial}{\partial u}\lfloor u \rceil \;=\;
\frac{\partial}{\partial u}\lfloor u \rfloor \;=\; 1,
\qquad\text{but}\qquad
\lfloor u \rceil \neq u,\;\; \lfloor u \rfloor \neq u
\quad\text{in the backward pass.}
\end{equation}
That is, the derivative of the rounding and floor operators is
approximated by~$1$, while their values are preserved.
Crucially, this means the rounded quantity
$r := \lfloor \tilde{x}/\mathit{ss}(\tilde{x}) \rceil$ does
\emph{not} collapse to $\tilde{x}/\mathit{ss}(\tilde{x})$ when
applying the chain rule; this distinction is what gives the
$m$-gradient a non-trivial form.

\paragraph{Setup.}
Let $r := \lfloor \tilde{x}/\mathit{ss}(\tilde{x}) \rceil$ and
$u := \mathit{ss}(\tilde{x})\cdot r$, so that $Q(\tilde{x}) = u$
on the unclipped range and
$Q(\tilde{x}) = q_{\max}^{(\mathrm{FP})}$ otherwise.
Recall that $\mathit{ss}(\tilde{x}) = 2^{\eta(\tilde{x})}$ with
$\eta$ as in Eq.~\ref{eq:appfp:exponent} and
$e_{\min} = 1 - b - m$ as in Eq.~\ref{eq:appfp:emin}.
The two regimes of $\eta$, (i.e., the normal regime
when $\eta = \lfloor \log_2 \tilde{x} \rfloor - m > e_{\min}$ and the
subnormal regime when $\eta = e_{\min}$), combined with whether the
input exceeds the saturating bound, partition the input range
into three cases: subnormal, normal-unclipped, and clipped
($\tilde{x} > q_{\max}^{(\mathrm{FP})}$).
Each yields a qualitatively different gradient and is kept
explicit below.
For $\tilde{x} > q_{\max}^{(\mathrm{FP})}$, the bit-width
gradients $\partial Q/\partial m$ and $\partial Q/\partial e$
remain non-trivial, since $q_{\max}^{(\mathrm{FP})}$ depends on
$m$ and $e$ through Eq.~\ref{eq:fp_range_rep}.
The negative branch follows by the symmetry
$Q(-\tilde{x}) = -Q(\tilde{x})$.
At $\tilde{x} = 0$, the gradient is defined as the limit of the
difference quotient from either side, taking the common value
when these limits agree, and zero otherwise; in our case the
bit-width gradients vanish since the rounding gap
$\tilde{x} - u \to 0$.

\paragraph{Gradient with respect to $m$.}
The mantissa bit-width $m$ enters $Q$ through $\eta$ (via the
$-m$ term in the normal regime and via $e_{\min}$ in the
subnormal regime) and, in the clipped regime, through
$q_{\max}^{(\mathrm{FP})}$.
For the unclipped range, $\partial \eta/\partial m = -1$ in both
the normal and subnormal regimes, so
\begin{equation}
\label{eq:appfp:dss_dm}
\frac{\partial \mathit{ss}}{\partial m}
\;=\;
-\ln(2)\cdot\mathit{ss}.
\end{equation}
Applying the STE to the round and using the chain rule on
$r = \lfloor \tilde{x}/\mathit{ss} \rceil$ at fixed $\tilde{x}$,
\begin{equation}
\frac{\partial r}{\partial m}
\;=\;
-\frac{\tilde{x}}{\mathit{ss}^{\,2}}\,
\frac{\partial \mathit{ss}}{\partial m}
\;=\;
\frac{\tilde{x}\,\ln(2)}{\mathit{ss}}.
\end{equation}
Combining via the product rule on $u = \mathit{ss}\cdot r$:
\begin{equation}
\frac{\partial Q}{\partial m}
\;=\;
\frac{\partial \mathit{ss}}{\partial m}\cdot r
\;+\; \mathit{ss}\cdot\frac{\partial r}{\partial m}
\;=\;
\ln(2)\cdot\bigl(\tilde{x} - u\bigr)
\qquad\text{for } |\tilde{x}| \in (0,\, q_{\max}^{(\mathrm{FP})}].
\end{equation}
This expression has the same form as TQT's threshold gradient
\cite[Eq.~7]{jain2020trained}, with $m$ substituted for the
log-threshold and the exclusion of $s$.

For $\tilde{x} > q_{\max}^{(\mathrm{FP})}$,
$Q = q_{\max}^{(\mathrm{FP})}$ and the $m$-derivative passes
through the clipping bound. Differentiating
Eq.~\ref{eq:fp_range},
\begin{equation}
\label{eq:appfp:dqmax_dm}
\frac{\partial q_{\max}^{(\mathrm{FP})}}{\partial m}
\;=\;
\ln(2)\cdot 2^{-m}\cdot 2^{\,2^{e} - b - 1}
\;=\;
\frac{\ln(2)}{2^{m+1} - 1}\cdot q_{\max}^{(\mathrm{FP})},
\end{equation}
a positive contribution that pushes $m$ upward to enlarge the
representable range and bring saturating elements inside it.
Combining the two regions yields the complete piecewise gradient
\begin{equation}
\label{eq:appfp:dQ_dm}
\frac{\partial Q}{\partial m}
\;=\;
\begin{cases}
\ln(2)\cdot\bigl(\tilde{x} - u\bigr)
   & \text{subnormal/normal regime, } |\tilde{x}| \in (0,\, q_{\max}^{(\mathrm{FP})}],\\[1.0ex]
\dfrac{\ln(2)}{2^{m+1} - 1}\cdot q_{\max}^{(\mathrm{FP})}
   & \tilde{x} > q_{\max}^{(\mathrm{FP})}.
\end{cases}
\end{equation}

\paragraph{Gradient with respect to $e$.}
The exponent bit-width $e$ enters $\eta$ only through
$b = 2^{e-1} - 1$, and only in the subnormal regime, since
$\lfloor\log_2 \tilde{x}\rfloor - m$ has is not a function of $e$.
Hence $\partial \eta/\partial e = 0$ in the normal regime and
$\partial \eta/\partial e = -\partial b/\partial e
   = -2^{e-1}\ln(2)$ in the subnormal regime.
Following the same chain-rule derivation as for $m$, with the
extra factor of $\partial \eta/\partial e$ propagating through
$\mathit{ss}$ and $r$:
\begin{equation}
\frac{\partial Q}{\partial e}
\;=\;
\begin{cases}
0
   & \text{normal regime},\\[0.5ex]
2^{e-1}\,(\ln 2)^{2}\cdot\bigl(\tilde{x} - u\bigr)
   & \text{subnormal regime } (\eta = e_{\min}),
\end{cases}
\qquad |\tilde{x}| \in (0,\, q_{\max}^{(\mathrm{FP})}].
\end{equation}

For $\tilde{x} > q_{\max}^{(\mathrm{FP})}$, the gradient passes
through $q_{\max}^{(\mathrm{FP})}$. With $b = 2^{e-1} - 1$ so
that $2^{e} - b - 1 = 2^{e-1}$, differentiating
Eq.~\ref{eq:fp_range} yields
\begin{equation}
\label{eq:appfp:dqmax_de}
\frac{\partial q_{\max}^{(\mathrm{FP})}}{\partial e}
\;=\;
2(\ln 2)^{2}\cdot 2^{e-1}\cdot 2^{\,2^{e} - b - 1}
   \bigl(1 - 2^{-m-1}\bigr)
\;=\;
(\ln 2)^{2}\cdot 2^{e-1}\cdot q_{\max}^{(\mathrm{FP})}.
\end{equation}
Combining the two regions yields the complete piecewise gradient
\begin{equation}
\label{eq:appfp:dQ_de}
\frac{\partial Q}{\partial e}
\;=\;
\begin{cases}
2^{e-1}\,(\ln 2)^{2}\cdot\bigl(\tilde{x} - u\bigr)
   & \text{subnormal regime } (\eta = e_{\min}),\\[1.0ex]
0
   & \text{normal regime, }
     |\tilde{x}| \in (0,\, q_{\max}^{(\mathrm{FP})}],\\[0.5ex]
(\ln 2)^{2}\cdot 2^{e-1}\cdot q_{\max}^{(\mathrm{FP})}
   & \tilde{x} > q_{\max}^{(\mathrm{FP})}.
\end{cases}
\end{equation}
The $e$-gradient is supported at \emph{both} extremes of the
input distribution: in the subnormal regime via $e_{\min}$, and
in the clipped regime via $q_{\max}^{(\mathrm{FP})}$,
while in the normal-unclipped regime the gradient contribution is zero.

\paragraph{Gradient with respect to $\beta$.}
The shared offset parameterization $m = 1 + \beta$,
$e = 2 + \beta$ (Section~\ref{subsec:cont_param}) gives
$\partial m/\partial \beta = \partial e/\partial \beta = 1$,
so the chain rule yields
\begin{equation}
\frac{\partial Q}{\partial \beta}
\;=\;
\frac{\partial Q}{\partial m}
\;+\;
\frac{\partial Q}{\partial e}.
\end{equation}
Substituting Eqs.~\ref{eq:appfp:dQ_dm} and~\ref{eq:appfp:dQ_de}
gives
\begin{equation}
\label{eq:appfp:dQ_dbeta}
\frac{\partial Q}{\partial \beta}
\;=\;
\begin{cases}
\ln(2)\,\bigl(1 + 2^{e-1}\ln 2\bigr)\cdot\bigl(\tilde{x} - u\bigr)
   & \text{subnormal regime } (\eta = e_{\min}),\\[1.0ex]
\ln(2)\cdot\bigl(\tilde{x} - u\bigr)
   & \text{normal regime, }
   |\tilde{x}| \in (0,\, q_{\max}^{(\mathrm{FP})}],\\[0.5ex]
\Bigl[\dfrac{\ln(2)}{2^{m+1} - 1}
       \,+\, (\ln 2)^{2}\,2^{e-1}\Bigr]
   \cdot q_{\max}^{(\mathrm{FP})}
   & \tilde{x} > q_{\max}^{(\mathrm{FP})}.
\end{cases}
\end{equation}
The $\beta$-gradient inherits the $m$-component's density across
all three regions, picking up additional $e$-component support
at the two extremes (subnormal and clipped).
Among the three format parameters $\{m, e, \beta\}$, only $m$
and $\beta$ carry a learning signal in the bulk
normal-unclipped regime where most of the input mass typically
lies, which motivates the $\beta$-only training scheme of the
main text.

\paragraph{Remarks.}
Three observations follow from
Eqs.~\ref{eq:appfp:dQ_dm}, \ref{eq:appfp:dQ_de}
and~\ref{eq:appfp:dQ_dbeta}, all of which are illustrated by
\HFigure{fig:appfp:bitwidth_gradients}.
First, the chain rule must be applied to the unsimplified
product $\mathit{ss}\cdot r$ rather than to its
forward-equivalent $\tilde{x}$; the STE preserves the
rounding gap $\tilde{x} - u$, which is the source of the
non-trivial $m$- and $e$-gradients in the unclipped range.
Second, the $m$-gradient is non-zero in all three regions
(subnormal, normal-unclipped, and clipped) and is the dominant
component of the $\beta$-gradient, as visible in
panels~\ref{fig:appfp:bitwidth_gradients:b}
and~\ref{fig:appfp:bitwidth_gradients:d} of
\HFigure{fig:appfp:bitwidth_gradients}.
Third, the $e$-gradient vanishes in the bulk normal-unclipped
regime and is supported at both extremes, i.e., below the
subnormal cutoff $2^{1-b}$ and above the clipping bound
$q_{\max}^{(\mathrm{FP})}$, so learning $e$ alone requires
sufficient mass at one or both extremes of the input
distribution, while learning the joint offset $\beta$ does not.
In practice these gradients are computed automatically by
autodifferentiation; we provide the closed forms here for
completeness, having verified them against PyTorch's
autograd~\cite{paszke2019pytorch}, confirmed by the overlap of
analytical and autograd curves in
\HFigure{fig:appfp:bitwidth_gradients}.

%
%
%

\pgfplotstableread[col sep=comma]%
  {images/gradient/bitwidth_gradients.csv}\bgdata

\pgfplotsset{
  bg axis/.style={
    width=0.95\linewidth,
    height=0.7\linewidth,
    xlabel={$x$},
    xmin=-10, xmax=10,
    grid=both,
    grid style={dashed, gray!20},
    every axis plot/.append style={line width=0.6pt},
    legend cell align=left,
    legend style={
      font=\small,
      draw=none,
      fill=none,
      inner sep=1pt,
    },
    label style={font=\small},
    tick label style={font=\small},
  },
}

\newcommand{\bgPanelA}{%
  \begin{tikzpicture}
    \begin{axis}[
        bg axis,
        ylabel={$x_q$},
        ymin=-8.5, ymax=8.5,
      ]
      \addplot[blue, thick] table[x=x, y=qx] {\bgdata};
      \draw[gray!50, dotted] (axis cs:7,-8.5)  -- (axis cs:7,8.5);
      \draw[gray!50, dotted] (axis cs:-7,-8.5) -- (axis cs:-7,8.5);
    \end{axis}
  \end{tikzpicture}%
}

\newcommand{\bgPanelB}{%
  \begin{tikzpicture}
    \begin{axis}[
        bg axis,
        ylabel={$\partial x_q / \partial m$},
        legend pos=north west,
      ]
      \addplot[black, dashed] table[x=x, y=dxq_dm] {\bgdata};
        \addlegendentry{analytical}
      \addplot[blue, dotted, line width=0.8pt]
        table[x=x, y=dxq_dm_brev] {\bgdata};
        \addlegendentry{autograd}
      \draw[gray!50, dotted]
        (axis cs:7,\pgfkeysvalueof{/pgfplots/ymin})
        -- (axis cs:7,\pgfkeysvalueof{/pgfplots/ymax});
      \draw[gray!50, dotted]
        (axis cs:-7,\pgfkeysvalueof{/pgfplots/ymin})
        -- (axis cs:-7,\pgfkeysvalueof{/pgfplots/ymax});
    \end{axis}
  \end{tikzpicture}%
}

\newcommand{\bgPanelC}{%
  \begin{tikzpicture}
    \begin{axis}[
        bg axis,
        ylabel={$\partial x_q / \partial e$},
        legend pos=north west,
      ]
      \addplot[black, dashed] table[x=x, y=dxq_de] {\bgdata};
        \addlegendentry{analytical}
      \addplot[blue, dotted, line width=0.8pt]
        table[x=x, y=dxq_de_brev] {\bgdata};
        \addlegendentry{autograd}
      \draw[gray!50, dotted]
        (axis cs:7,\pgfkeysvalueof{/pgfplots/ymin})
        -- (axis cs:7,\pgfkeysvalueof{/pgfplots/ymax});
      \draw[gray!50, dotted]
        (axis cs:-7,\pgfkeysvalueof{/pgfplots/ymin})
        -- (axis cs:-7,\pgfkeysvalueof{/pgfplots/ymax});
    \end{axis}
  \end{tikzpicture}%
}

\newcommand{\bgPanelD}{%
  \begin{tikzpicture}
    \begin{axis}[
        bg axis,
        ylabel={$\partial x_q / \partial \beta$},
        legend pos=north west,
      ]
      \addplot[black, dashed] table[x=x, y=dxq_dbeta] {\bgdata};
        \addlegendentry{analytical}
      \addplot[blue, dotted, line width=0.8pt]
        table[x=x, y=dxq_dbeta_brev] {\bgdata};
        \addlegendentry{autograd}
      \draw[gray!50, dotted]
        (axis cs:7,\pgfkeysvalueof{/pgfplots/ymin})
        -- (axis cs:7,\pgfkeysvalueof{/pgfplots/ymax});
      \draw[gray!50, dotted]
        (axis cs:-7,\pgfkeysvalueof{/pgfplots/ymin})
        -- (axis cs:-7,\pgfkeysvalueof{/pgfplots/ymax});
    \end{axis}
  \end{tikzpicture}%
}
\begin{figure}[t!]
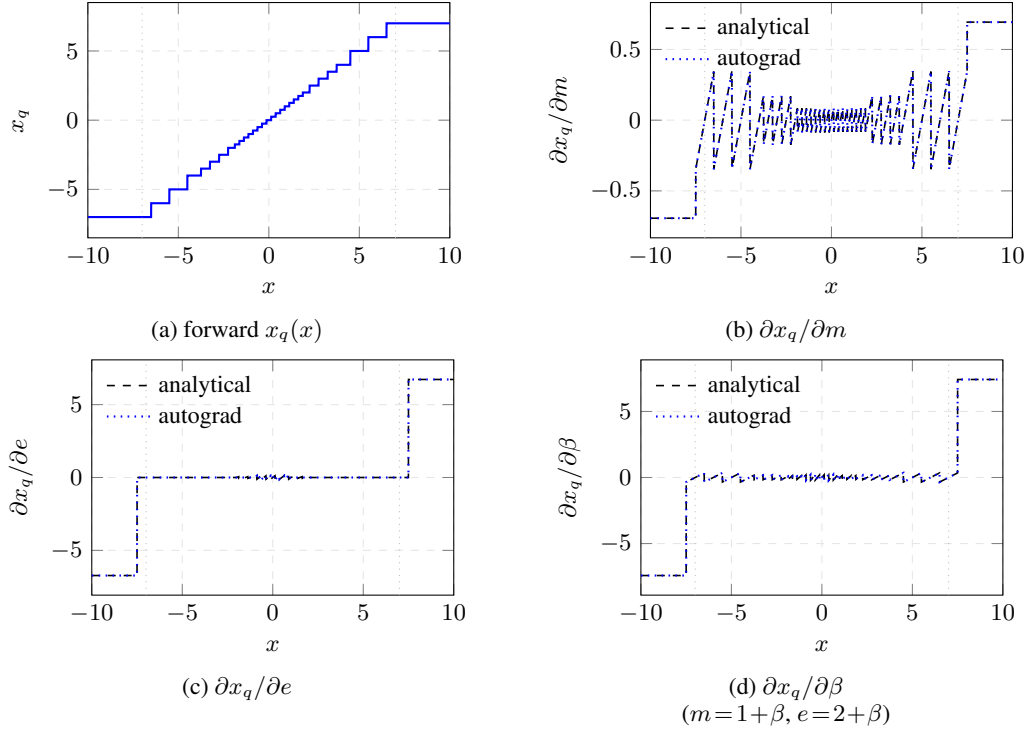

\centering
\begin{subfigure}[t]{0.48\textwidth}
  \centering
  \bgPanelA
  \caption{forward $x_q(x)$}
  \label{fig:appfp:bitwidth_gradients:a}
\end{subfigure}
\hfill
\begin{subfigure}[t]{0.48\textwidth}
  \centering
  \bgPanelB
  \caption{$\partial x_q / \partial m$}
  \label{fig:appfp:bitwidth_gradients:b}
\end{subfigure}
\\[1ex]
\begin{subfigure}[t]{0.48\textwidth}
  \centering
  \bgPanelC
  \caption{$\partial x_q / \partial e$}
  \label{fig:appfp:bitwidth_gradients:c}
\end{subfigure}
\hfill
\begin{subfigure}[t]{0.48\textwidth}
  \centering
  \bgPanelD
  \caption{$\partial x_q / \partial \beta$ \\
           ($m\!=\!1\!+\!\beta$, $e\!=\!2\!+\!\beta$)}
  \label{fig:appfp:bitwidth_gradients:d}
\end{subfigure}
\caption{Forward and bitwidth-gradient transfer curves of the
inner MX-FP quantizer at $e = 2$, $m = 2$.
The closed-form analytical gradients (black dashed) overlay the
Brevitas autograd reference (blue dotted) across the entire input
range.
Panels~\ref{fig:appfp:bitwidth_gradients:a}--\ref{fig:appfp:bitwidth_gradients:d}
show the forward $x_q(x)$ and the local gradients with respect
to $m$, $e$, and $\beta$, respectively.}
\label{fig:appfp:bitwidth_gradients}
\end{figure}

\section{Hyper-Parameters}\label{app:sec:hyperparam}
\subsection{Temperature Annealing}
We employ a temperature-based sigmoid function to switch from a linear-like approximation
to a step-like one during training.
For the scheduling of the temperature,
we employ an exponential function between the values of $8$ and $400$.
These values were picked because they best represent the linear-like and the step-like functions.

As already mentioned, we also employ a warm-up phase during which the temperature value
is kept constant so that the sigmoid is its linear-like phase.
After a fixed fraction of calibration steps, $T_{\mathrm{ratio}}$,
the exponential schedule starts.
For our experiments we set $T_{\mathrm{ratio}} = 60\%$.
We observed that our method is fairly robust to other values of $T_{\mathrm{ratio}}$,
although adapting this value to a specific model and training pipeline could potentially
yield better results.

\subsection{Bit-Width Cost Function}
When testing the simple scaling penalty defined in Eq.\ref{eq:bw_reg_simple},
we use 15 different values from $\lambda$ ranging from $1e^{-5}$ to $7$.
Similarly, in all the experiments with Eq.\ref{eq:bw_reg_target}, we fix $\lambda$ to $5$,
while testing for 17 different target bit-width $\bar{b}_{\mathrm{target}}$ between $4.1$ and $8$,
as mentioned in the Sec.\ref{sec:exp_setting}.

\subsection{Training Parameters}
We follow a similar set-up employed in SpinQuant~\cite{liu2024spinquant},
using the same hyper-parameters for the optimization of the rotation matrices.
In particular, we use CayleySGD optimizer with a learning rate of $1.5$.
We merge the Hadamard rotations whenever possible; these are then optimized during training.
For all layers where they cannot be merged, we leave them as standalone, static rotations.
The main difference is the number of training steps and training samples used,
which has been extended to $400$ and $3200$ respectively.

For the bit-width optimization, we employ SGD as optimized with a learning rate of $1$.

\section{Ablation Studies}\label{app:sec:ablation}

\begin{figure}[h!]
    \input{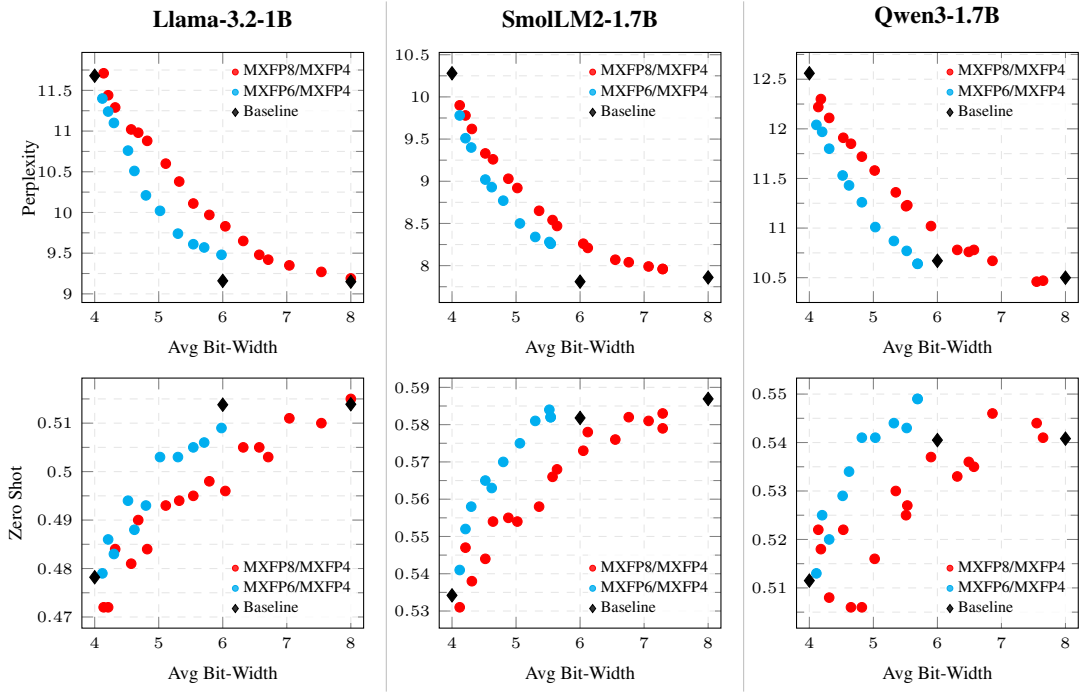}
    \pgfplotstableread[col sep=comma]{\csvAone}\dataAone
\pgfplotstableread[col sep=comma]{\csvBone}\dataBone
\pgfplotstableread[col sep=comma]{\csvCone}\dataCone

\ifshowSeriesB
  \pgfplotstableread[col sep=comma]{\csvAtwo}\dataAtwo
  \pgfplotstableread[col sep=comma]{\csvBtwo}\dataBtwo
  \pgfplotstableread[col sep=comma]{\csvCtwo}\dataCtwo
\fi

\pgfplotstableread[col sep=comma]{\refCsvA}\refDataA
\pgfplotstableread[col sep=comma]{\refCsvB}\refDataB
\pgfplotstableread[col sep=comma]{\refCsvC}\refDataC

\begin{tikzpicture}
  \begin{groupplot}[
      group style={
        group size=3 by 2,
        horizontal sep=1.0cm,
        vertical sep=1.0cm,
      },
      width=0.38\textwidth, height=0.35\textwidth,
      xmin=3.8, xmax=8.2,
      grid=both,
      grid style={dashed, gray!20},
      every axis plot/.append style={only marks, mark=*, mark size=1.8pt},
      legend cell align=left,
      legend style={
        font=\tiny,
        row sep=-2pt,
        draw=none,
        fill=white,
        fill opacity=0.8,
        text opacity=1,
        inner sep=0.5pt,
        legend image post style={scale=0.6},
      },
      title style={font=\small\bfseries},
      label style={font=\scriptsize},
      tick label style={font=\tiny},
      ylabel style={at={(-0.12,0.5)}},
      ytick distance=0.5,          
      minor y tick num=1,
    ]


    \nextgroupplot[
      title={\modelTitleA},
      xlabel={\xaxislabel},
      ylabel={Perplexity},
      legend style={at={(0.98,0.98)}, anchor=north east},
    ]
    \addplot[colorA] table[x=bitwidth, y=ppl] {\dataAone};
      \addlegendentry{\legendA}
    \ifshowSeriesB
      \addplot[colorB] table[x=bitwidth, y=ppl] {\dataAtwo};
        \addlegendentry{\legendB}
    \fi
    \addplot[black, mark=diamond*, mark size=2.5pt]
      table[x=bitwidth, y=ppl] {\refDataA};
      \addlegendentry{\legendRef}

    \nextgroupplot[
      title={\modelTitleB},
      xlabel={\xaxislabel},
      legend style={at={(0.98,0.98)}, anchor=north east},
    ]
    \addplot[colorA] table[x=bitwidth, y=ppl] {\dataBone};
      \addlegendentry{\legendA}
    \ifshowSeriesB
      \addplot[colorB] table[x=bitwidth, y=ppl] {\dataBtwo};
        \addlegendentry{\legendB}
    \fi
    \addplot[black, mark=diamond*, mark size=2.5pt]
      table[x=bitwidth, y=ppl] {\refDataB};
      \addlegendentry{\legendRef}

    \nextgroupplot[
      title={\modelTitleC},
      xlabel={\xaxislabel},
      legend style={at={(0.98,0.98)}, anchor=north east},
    ]
    \addplot[colorA] table[x=bitwidth, y=ppl] {\dataCone};
      \addlegendentry{\legendA}
    \ifshowSeriesB
      \addplot[colorB] table[x=bitwidth, y=ppl] {\dataCtwo};
        \addlegendentry{\legendB}
    \fi
    \addplot[black, mark=diamond*, mark size=2.5pt]
      table[x=bitwidth, y=ppl] {\refDataC};
      \addlegendentry{\legendRef}


    \nextgroupplot[
      xlabel={\xaxislabel},
      ylabel={Zero Shot},
      ylabel style={at={(-0.18,0.5)}},
      ytick distance=0.01,
      minor y tick num=1,
      legend style={at={(0.98,0.02)}, anchor=south east},
    ]
    \addplot[colorA] table[x=bitwidth, y=zs] {\dataAone};
      \addlegendentry{\legendA}
    \ifshowSeriesB
      \addplot[colorB] table[x=bitwidth, y=zs] {\dataAtwo};
        \addlegendentry{\legendB}
    \fi
    \addplot[black, mark=diamond*, mark size=2.5pt]
      table[x=bitwidth, y=zs] {\refDataA};
      \addlegendentry{\legendRef}

    \nextgroupplot[
      xlabel={\xaxislabel},
      ytick distance=0.01,
      minor y tick num=1,
      legend style={at={(0.98,0.02)}, anchor=south east},
    ]
    \addplot[colorA] table[x=bitwidth, y=zs] {\dataBone};
      \addlegendentry{\legendA}
    \ifshowSeriesB
      \addplot[colorB] table[x=bitwidth, y=zs] {\dataBtwo};
        \addlegendentry{\legendB}
    \fi
    \addplot[black, mark=diamond*, mark size=2.5pt]
      table[x=bitwidth, y=zs] {\refDataB};
      \addlegendentry{\legendRef}

    \nextgroupplot[
      xlabel={\xaxislabel},
      ytick distance=0.01,
      minor y tick num=1,
      legend style={at={(0.98,0.02)}, anchor=south east},
    ]
    \addplot[colorA] table[x=bitwidth, y=zs] {\dataCone};
      \addlegendentry{\legendA}
    \ifshowSeriesB
      \addplot[colorB] table[x=bitwidth, y=zs] {\dataCtwo};
        \addlegendentry{\legendB}
    \fi
    \addplot[black, mark=diamond*, mark size=2.5pt]
      table[x=bitwidth, y=zs] {\refDataC};
      \addlegendentry{\legendRef}

  \end{groupplot}

  \coordinate (sep12top) at ($ (group c1r1.east)!0.3!(group c2r1.west) $);
  \coordinate (sep12bot) at ($ (group c1r2.east)!0.3!(group c2r2.west) $);
  \coordinate (sep23top) at ($ (group c2r1.east)!0.3!(group c3r1.west) $);
  \coordinate (sep23bot) at ($ (group c2r2.east)!0.3!(group c3r2.west) $);
  \draw[gray!50, thin]
    (sep12top |- group c1r1.outer north) -- (sep12bot |- group c1r2.outer south);
  \draw[gray!50, thin]
    (sep23top |- group c2r1.outer north) -- (sep23bot |- group c2r2.outer south);
\end{tikzpicture}
\caption{Comparison of the MXFP8/MXFP4 mixed precision quantization and the
    MXFP6/MXFP4 ones.
    Red: MXFP8/MXFP4 configurations.
    Cyan: MXFP6/MXFP4 configurations.
    Black diamonds: homogeneous MXFP baselines.
    Top row: perplexity
    (lower is better).
    Bottom row: average zero-shot accuracy
    (higher is better).
    The MXFP6/MXFP4 configurations are more \textit{bit-width efficient},
    managing to reach better accuracy for the same average bit-width.}
\label{fig:mxfp8_mxfp6}
\end{figure}

\paragraph{Mixed Precision Results}
In \HFigure{fig:mxfp8_mxfp6} we represent the full set of experiments
where we compare MXFP8/MXFP4 mixed precision configurations to MXFP6/MXFP4 ones.
Although both configurations perform well,
the MXFP6/MXFP4 configurations are able to achieve better perplexity and
average zero-shot results for similar average bit-widths.

\paragraph{Continuous vs.\ discrete bit-width
representation.}
Several prior approaches store the bit-width as a
continuous variable but discretize it during each
forward pass,
typically through a rounding operator with an STE
in the backward
pass~\cite{bengio2013estimating}.
While effective for integer quantization where all
bit-widths in a range such as $[2, 8]$ are admissible,
this strategy may be less suitable for MXFP8--MXFP4
quantization.
The gap between FP8 and FP4 is comparatively large,
and abrupt format switches during calibration can
destabilize the training process.
The continuous representation used in \OurAlgo allows
the effective format to evolve gradually,
with the temperature-based annealing of
Section~\ref{subsec:temp_reg} managing the transition
to discrete inference-time values.

\begin{figure}[h!]
    \input{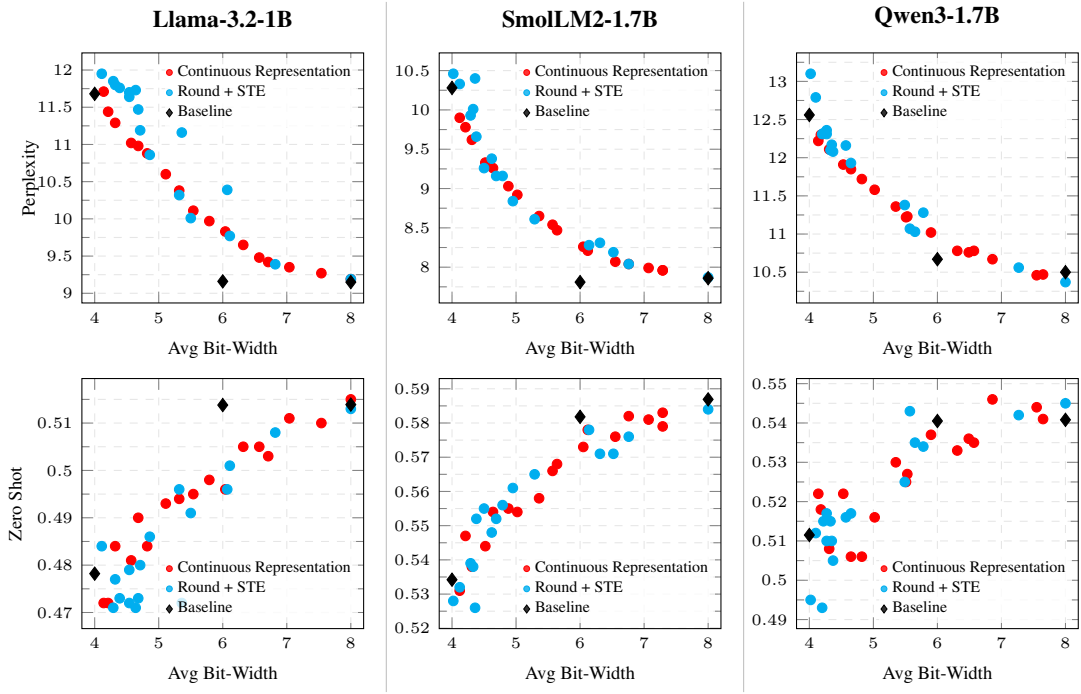}
    \pgfplotstableread[col sep=comma]{\csvAone}\dataAone
\pgfplotstableread[col sep=comma]{\csvBone}\dataBone
\pgfplotstableread[col sep=comma]{\csvCone}\dataCone

\ifshowSeriesB
  \pgfplotstableread[col sep=comma]{\csvAtwo}\dataAtwo
  \pgfplotstableread[col sep=comma]{\csvBtwo}\dataBtwo
  \pgfplotstableread[col sep=comma]{\csvCtwo}\dataCtwo
\fi

\pgfplotstableread[col sep=comma]{\refCsvA}\refDataA
\pgfplotstableread[col sep=comma]{\refCsvB}\refDataB
\pgfplotstableread[col sep=comma]{\refCsvC}\refDataC

\begin{tikzpicture}
  \begin{groupplot}[
      group style={
        group size=3 by 2,
        horizontal sep=1.0cm,
        vertical sep=1.0cm,
      },
      width=0.38\textwidth, height=0.35\textwidth,
      xmin=3.8, xmax=8.2,
      grid=both,
      grid style={dashed, gray!20},
      every axis plot/.append style={only marks, mark=*, mark size=1.8pt},
      legend cell align=left,
      legend style={
        font=\tiny,
        row sep=-2pt,
        draw=none,
        fill=white,
        fill opacity=0.8,
        text opacity=1,
        inner sep=0.5pt,
        legend image post style={scale=0.6},
      },
      title style={font=\small\bfseries},
      label style={font=\scriptsize},
      tick label style={font=\tiny},
      ylabel style={at={(-0.12,0.5)}},
      ytick distance=0.5,          
      minor y tick num=1,
    ]


    \nextgroupplot[
      title={\modelTitleA},
      xlabel={\xaxislabel},
      ylabel={Perplexity},
      legend style={at={(0.98,0.98)}, anchor=north east},
    ]
    \addplot[colorA] table[x=bitwidth, y=ppl] {\dataAone};
      \addlegendentry{\legendA}
    \ifshowSeriesB
      \addplot[colorB] table[x=bitwidth, y=ppl] {\dataAtwo};
        \addlegendentry{\legendB}
    \fi
    \addplot[black, mark=diamond*, mark size=2.5pt]
      table[x=bitwidth, y=ppl] {\refDataA};
      \addlegendentry{\legendRef}

    \nextgroupplot[
      title={\modelTitleB},
      xlabel={\xaxislabel},
      legend style={at={(0.98,0.98)}, anchor=north east},
    ]
    \addplot[colorA] table[x=bitwidth, y=ppl] {\dataBone};
      \addlegendentry{\legendA}
    \ifshowSeriesB
      \addplot[colorB] table[x=bitwidth, y=ppl] {\dataBtwo};
        \addlegendentry{\legendB}
    \fi
    \addplot[black, mark=diamond*, mark size=2.5pt]
      table[x=bitwidth, y=ppl] {\refDataB};
      \addlegendentry{\legendRef}

    \nextgroupplot[
      title={\modelTitleC},
      xlabel={\xaxislabel},
      legend style={at={(0.98,0.98)}, anchor=north east},
    ]
    \addplot[colorA] table[x=bitwidth, y=ppl] {\dataCone};
      \addlegendentry{\legendA}
    \ifshowSeriesB
      \addplot[colorB] table[x=bitwidth, y=ppl] {\dataCtwo};
        \addlegendentry{\legendB}
    \fi
    \addplot[black, mark=diamond*, mark size=2.5pt]
      table[x=bitwidth, y=ppl] {\refDataC};
      \addlegendentry{\legendRef}


    \nextgroupplot[
      xlabel={\xaxislabel},
      ylabel={Zero Shot},
      ylabel style={at={(-0.18,0.5)}},
      ytick distance=0.01,
      minor y tick num=1,
      legend style={at={(0.98,0.02)}, anchor=south east},
    ]
    \addplot[colorA] table[x=bitwidth, y=zs] {\dataAone};
      \addlegendentry{\legendA}
    \ifshowSeriesB
      \addplot[colorB] table[x=bitwidth, y=zs] {\dataAtwo};
        \addlegendentry{\legendB}
    \fi
    \addplot[black, mark=diamond*, mark size=2.5pt]
      table[x=bitwidth, y=zs] {\refDataA};
      \addlegendentry{\legendRef}

    \nextgroupplot[
      xlabel={\xaxislabel},
      ytick distance=0.01,
      minor y tick num=1,
      legend style={at={(0.98,0.02)}, anchor=south east},
    ]
    \addplot[colorA] table[x=bitwidth, y=zs] {\dataBone};
      \addlegendentry{\legendA}
    \ifshowSeriesB
      \addplot[colorB] table[x=bitwidth, y=zs] {\dataBtwo};
        \addlegendentry{\legendB}
    \fi
    \addplot[black, mark=diamond*, mark size=2.5pt]
      table[x=bitwidth, y=zs] {\refDataB};
      \addlegendentry{\legendRef}

    \nextgroupplot[
      xlabel={\xaxislabel},
      ytick distance=0.01,
      minor y tick num=1,
      legend style={at={(0.98,0.02)}, anchor=south east},
    ]
    \addplot[colorA] table[x=bitwidth, y=zs] {\dataCone};
      \addlegendentry{\legendA}
    \ifshowSeriesB
      \addplot[colorB] table[x=bitwidth, y=zs] {\dataCtwo};
        \addlegendentry{\legendB}
    \fi
    \addplot[black, mark=diamond*, mark size=2.5pt]
      table[x=bitwidth, y=zs] {\refDataC};
      \addlegendentry{\legendRef}

  \end{groupplot}

  \coordinate (sep12top) at ($ (group c1r1.east)!0.3!(group c2r1.west) $);
  \coordinate (sep12bot) at ($ (group c1r2.east)!0.3!(group c2r2.west) $);
  \coordinate (sep23top) at ($ (group c2r1.east)!0.3!(group c3r1.west) $);
  \coordinate (sep23bot) at ($ (group c2r2.east)!0.3!(group c3r2.west) $);
  \draw[gray!50, thin]
    (sep12top |- group c1r1.outer north) -- (sep12bot |- group c1r2.outer south);
  \draw[gray!50, thin]
    (sep23top |- group c2r1.outer north) -- (sep23bot |- group c2r2.outer south);
\end{tikzpicture}
\caption{Comparison between continuous forward-pass
    bit-width learning and a discretized forward pass
    with rounding (ROUND+STE).
    Red: continuous representation.
    Cyan: ROUND+STE.
    Black diamonds: homogeneous MXFP baselines.
    Top row: perplexity
    (lower is better).
    Bottom row: average zero-shot accuracy
    (higher is better).
    The continuous representation consistently
    achieves lower perplexity at a given average
    bit-width.}
\label{fig:continuous_vs_round}
\end{figure}

As shown in \HFigure{fig:continuous_vs_round},
the continuous representation consistently achieves
better perplexity and zero-shot than the ROUND+STE alternative
across all three models.
Even when they are comparable,
the results show that the ROUND+STE approach
struggles more in matching the desired target bit-width,
thus signaling a reliability issue compared to
the continuous representation format.

We believe that these results support the hypothesis that avoiding
abrupt format switches during the forward pass leads
to a more stable optimization trajectory.

\section{Results on Larger Models}\label{app:sec:big_models}
To verify that \OurAlgo scales beyond the models considered in the main text,
we evaluate it on four additional, larger models:
Llama~3.2~3B~\cite{grattafiori2024llama3},
Llama~3.1~8B~\cite{grattafiori2024llama3},
Qwen3~4B~\cite{yang2025qwen3},
and Qwen3~8B~\cite{yang2025qwen3}.
We follow the same evaluation protocol described in Section~\ref{sec:exp_setting},
reporting WikiText-2 perplexity and the average zero-shot accuracy over the four
reasoning benchmarks, and we compare the MXFP8/MXFP4 and MXFP6/MXFP4
mixed-precision configurations against the homogeneous MXFP baselines.

For these larger models we had to adapt two of the training hyper-parameters
with respect to the setup of Appendix~\ref{app:sec:hyperparam}.
Specifically, we increased the number of training steps to $800$
(from the $400$ used for the smaller models),
and we increased the learning rate of the bit-width optimization to $2$
(from the value of $1$ used previously).
We noticed that without these changes,
the bit-width parameters struggles to reach the desired target,
especially for higher target bit-widths.
All the remaining hyper-parameters are kept unchanged.

\begin{figure*}[h!]
    \input{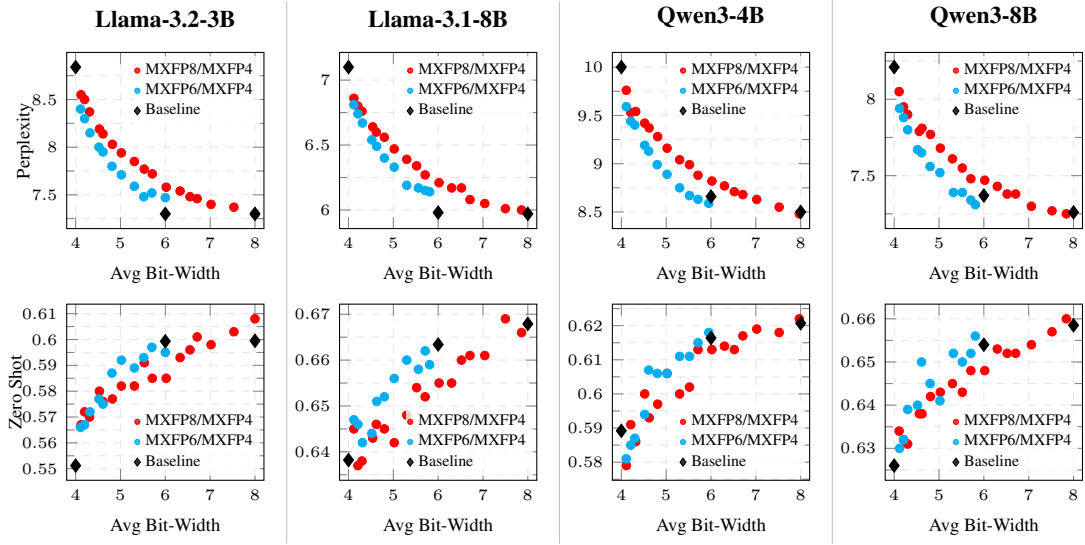}
%
%
%


\pgfplotstableread[col sep=comma]{\csvAone}\dataAone
\pgfplotstableread[col sep=comma]{\csvBone}\dataBone
\pgfplotstableread[col sep=comma]{\csvCone}\dataCone
\pgfplotstableread[col sep=comma]{\csvDone}\dataDone

\ifshowSeriesB
  \pgfplotstableread[col sep=comma]{\csvAtwo}\dataAtwo
  \pgfplotstableread[col sep=comma]{\csvBtwo}\dataBtwo
  \pgfplotstableread[col sep=comma]{\csvCtwo}\dataCtwo
  \pgfplotstableread[col sep=comma]{\csvDtwo}\dataDtwo
\fi

\pgfplotstableread[col sep=comma]{\refCsvA}\refDataA
\pgfplotstableread[col sep=comma]{\refCsvB}\refDataB
\pgfplotstableread[col sep=comma]{\refCsvC}\refDataC
\pgfplotstableread[col sep=comma]{\refCsvD}\refDataD

\begin{tikzpicture}
  \begin{groupplot}[
      group style={
        group size=4 by 2,
        horizontal sep=1.0cm,
        vertical sep=1.0cm,
      },
      width=0.30\textwidth, height=0.28\textwidth,
      xmin=3.8, xmax=8.2,
      grid=both,
      grid style={dashed, gray!20},
      every axis plot/.append style={only marks, mark=*, mark size=1.6pt},
      legend cell align=left,
      legend style={
        font=\tiny,
        row sep=-2pt,
        draw=none,
        fill=white,
        fill opacity=0.8,
        text opacity=1,
        inner sep=0.5pt,
        legend image post style={scale=0.6},
      },
      title style={font=\small\bfseries},
      label style={font=\scriptsize},
      tick label style={font=\tiny},
      ylabel style={at={(-0.12,0.5)}},
      ytick distance=0.5,          
      minor y tick num=1,
    ]


    \nextgroupplot[
      title={\modelTitleA},
      xlabel={\xaxislabel},
      ylabel={Perplexity},
      legend style={at={(0.98,0.98)}, anchor=north east},
    ]
    \addplot[colorA] table[x=bitwidth, y=ppl] {\dataAone};
      \addlegendentry{\legendA}
    \ifshowSeriesB
      \addplot[colorB] table[x=bitwidth, y=ppl] {\dataAtwo};
        \addlegendentry{\legendB}
    \fi
    \addplot[black, mark=diamond*, mark size=2.5pt]
      table[x=bitwidth, y=ppl] {\refDataA};
      \addlegendentry{\legendRef}

    \nextgroupplot[
      title={\modelTitleB},
      xlabel={\xaxislabel},
      legend style={at={(0.98,0.98)}, anchor=north east},
    ]
    \addplot[colorA] table[x=bitwidth, y=ppl] {\dataBone};
      \addlegendentry{\legendA}
    \ifshowSeriesB
      \addplot[colorB] table[x=bitwidth, y=ppl] {\dataBtwo};
        \addlegendentry{\legendB}
    \fi
    \addplot[black, mark=diamond*, mark size=2.5pt]
      table[x=bitwidth, y=ppl] {\refDataB};
      \addlegendentry{\legendRef}

    \nextgroupplot[
      title={\modelTitleC},
      xlabel={\xaxislabel},
      legend style={at={(0.98,0.98)}, anchor=north east},
    ]
    \addplot[colorA] table[x=bitwidth, y=ppl] {\dataCone};
      \addlegendentry{\legendA}
    \ifshowSeriesB
      \addplot[colorB] table[x=bitwidth, y=ppl] {\dataCtwo};
        \addlegendentry{\legendB}
    \fi
    \addplot[black, mark=diamond*, mark size=2.5pt]
      table[x=bitwidth, y=ppl] {\refDataC};
      \addlegendentry{\legendRef}

    \nextgroupplot[
      title={\modelTitleD},
      xlabel={\xaxislabel},
      legend style={at={(0.98,0.98)}, anchor=north east},
    ]
    \addplot[colorA] table[x=bitwidth, y=ppl] {\dataDone};
      \addlegendentry{\legendA}
    \ifshowSeriesB
      \addplot[colorB] table[x=bitwidth, y=ppl] {\dataDtwo};
        \addlegendentry{\legendB}
    \fi
    \addplot[black, mark=diamond*, mark size=2.5pt]
      table[x=bitwidth, y=ppl] {\refDataD};
      \addlegendentry{\legendRef}


    \nextgroupplot[
      xlabel={\xaxislabel},
      ylabel={Zero Shot},
      ylabel style={at={(-0.18,0.5)}},
      ytick distance=0.01,
      minor y tick num=1,
      legend style={at={(0.98,0.02)}, anchor=south east},
    ]
    \addplot[colorA] table[x=bitwidth, y=zs] {\dataAone};
      \addlegendentry{\legendA}
    \ifshowSeriesB
      \addplot[colorB] table[x=bitwidth, y=zs] {\dataAtwo};
        \addlegendentry{\legendB}
    \fi
    \addplot[black, mark=diamond*, mark size=2.5pt]
      table[x=bitwidth, y=zs] {\refDataA};
      \addlegendentry{\legendRef}

    \nextgroupplot[
      xlabel={\xaxislabel},
      ytick distance=0.01,
      minor y tick num=1,
      legend style={at={(0.98,0.02)}, anchor=south east},
    ]
    \addplot[colorA] table[x=bitwidth, y=zs] {\dataBone};
      \addlegendentry{\legendA}
    \ifshowSeriesB
      \addplot[colorB] table[x=bitwidth, y=zs] {\dataBtwo};
        \addlegendentry{\legendB}
    \fi
    \addplot[black, mark=diamond*, mark size=2.5pt]
      table[x=bitwidth, y=zs] {\refDataB};
      \addlegendentry{\legendRef}

    \nextgroupplot[
      xlabel={\xaxislabel},
      ytick distance=0.01,
      minor y tick num=1,
      legend style={at={(0.98,0.02)}, anchor=south east},
    ]
    \addplot[colorA] table[x=bitwidth, y=zs] {\dataCone};
      \addlegendentry{\legendA}
    \ifshowSeriesB
      \addplot[colorB] table[x=bitwidth, y=zs] {\dataCtwo};
        \addlegendentry{\legendB}
    \fi
    \addplot[black, mark=diamond*, mark size=2.5pt]
      table[x=bitwidth, y=zs] {\refDataC};
      \addlegendentry{\legendRef}

    \nextgroupplot[
      xlabel={\xaxislabel},
      ytick distance=0.01,
      minor y tick num=1,
      legend style={at={(0.98,0.02)}, anchor=south east},
    ]
    \addplot[colorA] table[x=bitwidth, y=zs] {\dataDone};
      \addlegendentry{\legendA}
    \ifshowSeriesB
      \addplot[colorB] table[x=bitwidth, y=zs] {\dataDtwo};
        \addlegendentry{\legendB}
    \fi
    \addplot[black, mark=diamond*, mark size=2.5pt]
      table[x=bitwidth, y=zs] {\refDataD};
      \addlegendentry{\legendRef}

  \end{groupplot}

  \coordinate (sep12top) at ($ (group c1r1.east)!0.3!(group c2r1.west) $);
  \coordinate (sep12bot) at ($ (group c1r2.east)!0.3!(group c2r2.west) $);
  \coordinate (sep23top) at ($ (group c2r1.east)!0.3!(group c3r1.west) $);
  \coordinate (sep23bot) at ($ (group c2r2.east)!0.3!(group c3r2.west) $);
  \coordinate (sep34top) at ($ (group c3r1.east)!0.3!(group c4r1.west) $);
  \coordinate (sep34bot) at ($ (group c3r2.east)!0.3!(group c4r2.west) $);
  \draw[gray!50, thin]
    (sep12top |- group c1r1.outer north) -- (sep12bot |- group c1r2.outer south);
  \draw[gray!50, thin]
    (sep23top |- group c2r1.outer north) -- (sep23bot |- group c2r2.outer south);
  \draw[gray!50, thin]
    (sep34top |- group c3r1.outer north) -- (sep34bot |- group c3r2.outer south);
\end{tikzpicture}
\caption{Mixed-precision quantization results on larger models
    (Llama-3.2-3B, Llama-3.1-8B, Qwen3-4B, Qwen3-8B).
    Red: MXFP8/MXFP4 configurations.
    Cyan: MXFP6/MXFP4 configurations.
    Black diamonds: homogeneous MXFP baselines.
    Top row: perplexity
    (lower is better).
    Bottom row: average zero-shot accuracy
    (higher is better).
    As with the smaller models, the MXFP6/MXFP4 configurations are more
    \textit{bit-width efficient}, reaching better accuracy for the same
    average bit-width.}
\label{fig:big_models}
\end{figure*}

\HFigure{fig:big_models} reports the full set of operating points
for all four models.
The trends observed for the smaller models carry over to this larger scale:
both mixed-precision configurations consistently improve over the MXFP4 baseline
as the average bit-width increases,
and the MXFP6/MXFP4 configurations remain more \textit{bit-width efficient},
reaching comparable or better perplexity and zero-shot accuracy than the
MXFP8/MXFP4 configurations at the same average bit-width.
This confirms that \OurAlgo generalizes to larger models without
substantial changes to the method.

One interesting point to mention is that, for Llama models,
the MXFP6/MXFP4 configurations seem to underperform compared to the MXFP6 baseline,
at least in terms of perplexity, whilst zero-shot performance remains comparable.
Conversely, for Qwen Models, \OurAlgo~ seems to perform better than the baseline.
We argue that this mostly signifies that,
although the training hypper-parameters we picked seem to perform generally well,
in some cases it might be necessary to adapt the training configuration to the specific model considered.

\section{High-Precision Layer Frequency}\label{app:sec:frequency}
To understand which layers \OurAlgo most consistently retains at high
precision, we compute, for each layer, the fraction of mixed-precision
configurations in which it is assigned the high-precision format,
thereby obtaining an empirical sensitivity ranking across operating points.
We report the $30$ most frequently high-precision layers, grouped by model
family: Tables~\ref{tab:freq_llama_mxfp8} and~\ref{tab:freq_llama_mxfp6}
cover the Llama family under the MXFP8/MXFP4 and MXFP6/MXFP4 configurations,
respectively; Tables~\ref{tab:freq_qwen_mxfp8} and~\ref{tab:freq_qwen_mxfp6}
cover the Qwen family; and Table~\ref{tab:freq_smollm2} reports both formats
for SmolLM2-1.7B.

A consistent pattern emerges across models and configurations:
\texttt{down\_proj} and \texttt{v\_proj} layers are most frequently assigned
the high-precision format, indicating that they are systematically more
sensitive to quantization than other layer types.

Beyond this shared tendency, the sensitivity ranking varies considerably
across quantization configurations, model families, and model sizes.
This variability highlights that quantization sensitivity is difficult to
predict in general: even minor architectural differences can substantially
alter the relative ordering of layer sensitivities, limiting the degree to
which rankings obtained on one model or configuration can be transferred to
another.

\begin{table}[t]
  \centering
  \footnotesize
  \setlength{\tabcolsep}{4pt}
  \caption{Top-30 layers most frequently kept at high precision in the
    MXFP8/MXFP4 mixed-precision experiments for the Llama family.
    ``Freq.'' is the fraction of configurations in which the layer was
    assigned the high-precision (MXFP8) format.}
  \label{tab:freq_llama_mxfp8}
  \begin{tabular}{r ll ll ll}
    \toprule
    & \multicolumn{2}{c}{Llama-3.2-1B}
    & \multicolumn{2}{c}{Llama-3.2-3B}
    & \multicolumn{2}{c}{Llama-3.1-8B} \\
    \cmidrule(lr){2-3}\cmidrule(lr){4-5}\cmidrule(lr){6-7}
    Rank & Layer & Freq. & Layer & Freq. & Layer & Freq. \\
    \midrule
    1  & \texttt{down\_proj\_1}  & 1.00 & \texttt{down\_proj\_1}  & 1.00 & \texttt{down\_proj\_1}  & 1.00 \\
    2  & \texttt{v\_proj\_1}     & 1.00 & \texttt{v\_proj\_2}     & 0.98 & \texttt{down\_proj\_31} & 1.00 \\
    3  & \texttt{v\_proj\_2}     & 1.00 & \texttt{v\_proj\_3}     & 0.98 & \texttt{v\_proj\_2}     & 1.00 \\
    4  & \texttt{down\_proj\_15} & 0.94 & \texttt{down\_proj\_27} & 0.96 & \texttt{v\_proj\_3}     & 0.98 \\
    5  & \texttt{v\_proj\_3}     & 0.94 & \texttt{v\_proj\_1}     & 0.96 & \texttt{v\_proj\_4}     & 0.98 \\
    6  & \texttt{down\_proj\_0}  & 0.88 & \texttt{v\_proj\_4}     & 0.94 & \texttt{v\_proj\_19}    & 0.96 \\
    7  & \texttt{v\_proj\_12}    & 0.82 & \texttt{v\_proj\_5}     & 0.94 & \texttt{v\_proj\_5}     & 0.94 \\
    8  & \texttt{v\_proj\_13}    & 0.82 & \texttt{v\_proj\_20}    & 0.92 & \texttt{up\_proj\_31}   & 0.92 \\
    9  & \texttt{v\_proj\_4}     & 0.82 & \texttt{v\_proj\_21}    & 0.92 & \texttt{v\_proj\_1}     & 0.92 \\
    10 & \texttt{v\_proj\_5}     & 0.82 & \texttt{v\_proj\_18}    & 0.90 & \texttt{v\_proj\_24}    & 0.92 \\
    11 & \texttt{v\_proj\_7}     & 0.82 & \texttt{v\_proj\_25}    & 0.90 & \texttt{v\_proj\_18}    & 0.90 \\
    12 & \texttt{v\_proj\_9}     & 0.82 & \texttt{v\_proj\_22}    & 0.88 & \texttt{v\_proj\_23}    & 0.88 \\
    13 & \texttt{down\_proj\_11} & 0.76 & \texttt{v\_proj\_24}    & 0.88 & \texttt{down\_proj\_0}  & 0.86 \\
    14 & \texttt{up\_proj\_9}    & 0.76 & \texttt{v\_proj\_16}    & 0.86 & \texttt{v\_proj\_6}     & 0.86 \\
    15 & \texttt{v\_proj\_14}    & 0.76 & \texttt{v\_proj\_19}    & 0.86 & \texttt{v\_proj\_0}     & 0.84 \\
    16 & \texttt{down\_proj\_10} & 0.71 & \texttt{v\_proj\_15}    & 0.84 & \texttt{v\_proj\_15}    & 0.84 \\
    17 & \texttt{gate\_proj\_15} & 0.71 & \texttt{v\_proj\_17}    & 0.84 & \texttt{v\_proj\_17}    & 0.84 \\
    18 & \texttt{up\_proj\_10}   & 0.71 & \texttt{down\_proj\_0}  & 0.82 & \texttt{v\_proj\_20}    & 0.84 \\
    19 & \texttt{up\_proj\_11}   & 0.71 & \texttt{v\_proj\_14}    & 0.82 & \texttt{v\_proj\_21}    & 0.84 \\
    20 & \texttt{up\_proj\_15}   & 0.71 & \texttt{v\_proj\_23}    & 0.82 & \texttt{v\_proj\_9}     & 0.84 \\
    21 & \texttt{v\_proj\_10}    & 0.71 & \texttt{v\_proj\_6}     & 0.82 & \texttt{v\_proj\_10}    & 0.82 \\
    22 & \texttt{v\_proj\_11}    & 0.71 & \texttt{v\_proj\_9}     & 0.78 & \texttt{v\_proj\_25}    & 0.82 \\
    23 & \texttt{v\_proj\_15}    & 0.71 & \texttt{v\_proj\_0}     & 0.75 & \texttt{v\_proj\_26}    & 0.82 \\
    24 & \texttt{down\_proj\_14} & 0.65 & \texttt{v\_proj\_10}    & 0.75 & \texttt{v\_proj\_22}    & 0.80 \\
    25 & \texttt{down\_proj\_9}  & 0.65 & \texttt{v\_proj\_11}    & 0.75 & \texttt{v\_proj\_13}    & 0.78 \\
    26 & \texttt{up\_proj\_0}    & 0.65 & \texttt{v\_proj\_7}     & 0.75 & \texttt{v\_proj\_28}    & 0.78 \\
    27 & \texttt{up\_proj\_2}    & 0.65 & \texttt{v\_proj\_8}     & 0.75 & \texttt{v\_proj\_29}    & 0.78 \\
    28 & \texttt{up\_proj\_4}    & 0.65 & \texttt{v\_proj\_12}    & 0.73 & \texttt{gate\_proj\_31} & 0.76 \\
    29 & \texttt{v\_proj\_0}     & 0.65 & \texttt{v\_proj\_26}    & 0.71 & \texttt{v\_proj\_11}    & 0.76 \\
    30 & \texttt{down\_proj\_4}  & 0.59 & \texttt{down\_proj\_14} & 0.69 & \texttt{v\_proj\_12}    & 0.76 \\
    \bottomrule
  \end{tabular}
\end{table}

\begin{table}[t]
  \centering
  \footnotesize
  \setlength{\tabcolsep}{4pt}
  \caption{Top-30 layers most frequently kept at high precision in the
    MXFP6/MXFP4 mixed-precision experiments for the Llama family.
    ``Freq.'' is the fraction of configurations in which the layer was
    assigned the high-precision (MXFP6) format.}
  \label{tab:freq_llama_mxfp6}
  \begin{tabular}{r ll ll ll}
    \toprule
    & \multicolumn{2}{c}{Llama-3.2-1B}
    & \multicolumn{2}{c}{Llama-3.2-3B}
    & \multicolumn{2}{c}{Llama-3.1-8B} \\
    \cmidrule(lr){2-3}\cmidrule(lr){4-5}\cmidrule(lr){6-7}
    Rank & Layer & Freq. & Layer & Freq. & Layer & Freq. \\
    \midrule
    1  & \texttt{down\_proj\_0}  & 1.00 & \texttt{down\_proj\_1}  & 0.94 & \texttt{down\_proj\_1}  & 0.96 \\
    2  & \texttt{down\_proj\_1}  & 1.00 & \texttt{v\_proj\_2}     & 0.94 & \texttt{down\_proj\_31} & 0.96 \\
    3  & \texttt{down\_proj\_15} & 1.00 & \texttt{v\_proj\_21}    & 0.94 & \texttt{v\_proj\_1}     & 0.96 \\
    4  & \texttt{v\_proj\_2}     & 1.00 & \texttt{v\_proj\_3}     & 0.94 & \texttt{v\_proj\_2}     & 0.96 \\
    5  & \texttt{v\_proj\_3}     & 1.00 & \texttt{v\_proj\_4}     & 0.94 & \texttt{v\_proj\_3}     & 0.96 \\
    6  & \texttt{v\_proj\_1}     & 0.91 & \texttt{down\_proj\_27} & 0.92 & \texttt{v\_proj\_4}     & 0.96 \\
    7  & \texttt{v\_proj\_10}    & 0.91 & \texttt{v\_proj\_1}     & 0.92 & \texttt{v\_proj\_5}     & 0.96 \\
    8  & \texttt{v\_proj\_13}    & 0.91 & \texttt{v\_proj\_17}    & 0.92 & \texttt{v\_proj\_18}    & 0.94 \\
    9  & \texttt{v\_proj\_14}    & 0.91 & \texttt{v\_proj\_18}    & 0.92 & \texttt{v\_proj\_19}    & 0.92 \\
    10 & \texttt{v\_proj\_9}     & 0.91 & \texttt{v\_proj\_22}    & 0.89 & \texttt{v\_proj\_20}    & 0.91 \\
    11 & \texttt{up\_proj\_4}    & 0.82 & \texttt{v\_proj\_5}     & 0.89 & \texttt{up\_proj\_31}   & 0.89 \\
    12 & \texttt{up\_proj\_8}    & 0.82 & \texttt{v\_proj\_20}    & 0.87 & \texttt{v\_proj\_0}     & 0.89 \\
    13 & \texttt{up\_proj\_9}    & 0.82 & \texttt{down\_proj\_0}  & 0.85 & \texttt{v\_proj\_17}    & 0.89 \\
    14 & \texttt{v\_proj\_11}    & 0.82 & \texttt{v\_proj\_16}    & 0.85 & \texttt{v\_proj\_23}    & 0.89 \\
    15 & \texttt{v\_proj\_12}    & 0.82 & \texttt{v\_proj\_6}     & 0.85 & \texttt{down\_proj\_0}  & 0.87 \\
    16 & \texttt{v\_proj\_4}     & 0.82 & \texttt{v\_proj\_19}    & 0.83 & \texttt{v\_proj\_15}    & 0.87 \\
    17 & \texttt{v\_proj\_5}     & 0.82 & \texttt{v\_proj\_25}    & 0.83 & \texttt{v\_proj\_24}    & 0.87 \\
    18 & \texttt{v\_proj\_7}     & 0.82 & \texttt{v\_proj\_0}     & 0.81 & \texttt{v\_proj\_21}    & 0.85 \\
    19 & \texttt{down\_proj\_10} & 0.73 & \texttt{v\_proj\_14}    & 0.79 & \texttt{v\_proj\_25}    & 0.85 \\
    20 & \texttt{down\_proj\_11} & 0.73 & \texttt{v\_proj\_15}    & 0.79 & \texttt{v\_proj\_6}     & 0.85 \\
    21 & \texttt{up\_proj\_11}   & 0.73 & \texttt{v\_proj\_23}    & 0.77 & \texttt{v\_proj\_29}    & 0.81 \\
    22 & \texttt{up\_proj\_15}   & 0.73 & \texttt{v\_proj\_9}     & 0.77 & \texttt{v\_proj\_9}     & 0.81 \\
    23 & \texttt{up\_proj\_3}    & 0.73 & \texttt{v\_proj\_24}    & 0.75 & \texttt{v\_proj\_22}    & 0.79 \\
    24 & \texttt{up\_proj\_5}    & 0.73 & \texttt{v\_proj\_7}     & 0.75 & \texttt{v\_proj\_26}    & 0.79 \\
    25 & \texttt{down\_proj\_2}  & 0.64 & \texttt{v\_proj\_26}    & 0.72 & \texttt{v\_proj\_10}    & 0.77 \\
    26 & \texttt{down\_proj\_3}  & 0.64 & \texttt{v\_proj\_11}    & 0.70 & \texttt{v\_proj\_16}    & 0.77 \\
    27 & \texttt{down\_proj\_9}  & 0.64 & \texttt{v\_proj\_10}    & 0.68 & \texttt{v\_proj\_30}    & 0.77 \\
    28 & \texttt{up\_proj\_0}    & 0.64 & \texttt{v\_proj\_12}    & 0.66 & \texttt{v\_proj\_7}     & 0.77 \\
    29 & \texttt{up\_proj\_10}   & 0.64 & \texttt{v\_proj\_8}     & 0.66 & \texttt{gate\_proj\_31} & 0.75 \\
    30 & \texttt{up\_proj\_2}    & 0.64 & \texttt{gate\_proj\_27} & 0.64 & \texttt{v\_proj\_27}    & 0.75 \\
    \bottomrule
  \end{tabular}
\end{table}

\begin{table}[t]
  \centering
  \footnotesize
  \setlength{\tabcolsep}{4pt}
  \caption{Top-30 layers most frequently kept at high precision in the
    MXFP8/MXFP4 mixed-precision experiments for the Qwen family.
    ``Freq.'' is the fraction of configurations in which the layer was
    assigned the high-precision (MXFP8) format.}
  \label{tab:freq_qwen_mxfp8}
  \begin{tabular}{r ll ll ll}
    \toprule
    & \multicolumn{2}{c}{Qwen3-1.7B}
    & \multicolumn{2}{c}{Qwen3-4B}
    & \multicolumn{2}{c}{Qwen3-8B} \\
    \cmidrule(lr){2-3}\cmidrule(lr){4-5}\cmidrule(lr){6-7}
    Rank & Layer & Freq. & Layer & Freq. & Layer & Freq. \\
    \midrule
    1  & \texttt{down\_proj\_2}  & 1.00 & \texttt{down\_proj\_35} & 1.00 & \texttt{down\_proj\_35} & 0.96 \\
    2  & \texttt{down\_proj\_27} & 1.00 & \texttt{down\_proj\_6}  & 1.00 & \texttt{down\_proj\_6}  & 0.96 \\
    3  & \texttt{v\_proj\_25}    & 1.00 & \texttt{v\_proj\_34}    & 0.98 & \texttt{up\_proj\_35}   & 0.96 \\
    4  & \texttt{v\_proj\_27}    & 1.00 & \texttt{v\_proj\_26}    & 0.96 & \texttt{v\_proj\_30}    & 0.96 \\
    5  & \texttt{up\_proj\_27}   & 0.94 & \texttt{v\_proj\_33}    & 0.96 & \texttt{v\_proj\_33}    & 0.96 \\
    6  & \texttt{v\_proj\_24}    & 0.94 & \texttt{gate\_proj\_35} & 0.92 & \texttt{v\_proj\_34}    & 0.96 \\
    7  & \texttt{v\_proj\_26}    & 0.94 & \texttt{v\_proj\_29}    & 0.92 & \texttt{v\_proj\_29}    & 0.94 \\
    8  & \texttt{v\_proj\_3}     & 0.94 & \texttt{v\_proj\_31}    & 0.92 & \texttt{v\_proj\_16}    & 0.92 \\
    9  & \texttt{v\_proj\_9}     & 0.94 & \texttt{up\_proj\_35}   & 0.90 & \texttt{v\_proj\_26}    & 0.92 \\
    10 & \texttt{up\_proj\_26}   & 0.88 & \texttt{v\_proj\_22}    & 0.90 & \texttt{v\_proj\_27}    & 0.92 \\
    11 & \texttt{v\_proj\_0}     & 0.88 & \texttt{v\_proj\_30}    & 0.90 & \texttt{v\_proj\_10}    & 0.91 \\
    12 & \texttt{v\_proj\_15}    & 0.88 & \texttt{v\_proj\_25}    & 0.88 & \texttt{v\_proj\_22}    & 0.91 \\
    13 & \texttt{v\_proj\_19}    & 0.88 & \texttt{v\_proj\_32}    & 0.88 & \texttt{gate\_proj\_35} & 0.89 \\
    14 & \texttt{gate\_proj\_27} & 0.82 & \texttt{v\_proj\_27}    & 0.86 & \texttt{v\_proj\_19}    & 0.87 \\
    15 & \texttt{o\_proj\_0}     & 0.82 & \texttt{v\_proj\_35}    & 0.86 & \texttt{v\_proj\_24}    & 0.87 \\
    16 & \texttt{v\_proj\_20}    & 0.82 & \texttt{down\_proj\_16} & 0.84 & \texttt{v\_proj\_32}    & 0.87 \\
    17 & \texttt{v\_proj\_21}    & 0.82 & \texttt{v\_proj\_10}    & 0.84 & \texttt{v\_proj\_35}    & 0.87 \\
    18 & \texttt{v\_proj\_23}    & 0.82 & \texttt{k\_proj\_33}    & 0.82 & \texttt{down\_proj\_16} & 0.85 \\
    19 & \texttt{v\_proj\_4}     & 0.82 & \texttt{v\_proj\_16}    & 0.82 & \texttt{v\_proj\_8}     & 0.85 \\
    20 & \texttt{v\_proj\_5}     & 0.82 & \texttt{v\_proj\_17}    & 0.82 & \texttt{v\_proj\_9}     & 0.85 \\
    21 & \texttt{v\_proj\_8}     & 0.82 & \texttt{v\_proj\_19}    & 0.82 & \texttt{v\_proj\_17}    & 0.83 \\
    22 & \texttt{down\_proj\_26} & 0.76 & \texttt{v\_proj\_23}    & 0.82 & \texttt{up\_proj\_34}   & 0.81 \\
    23 & \texttt{up\_proj\_5}    & 0.76 & \texttt{v\_proj\_8}     & 0.82 & \texttt{v\_proj\_15}    & 0.81 \\
    24 & \texttt{v\_proj\_16}    & 0.76 & \texttt{v\_proj\_9}     & 0.82 & \texttt{up\_proj\_15}   & 0.79 \\
    25 & \texttt{v\_proj\_17}    & 0.76 & \texttt{up\_proj\_18}   & 0.80 & \texttt{up\_proj\_16}   & 0.77 \\
    26 & \texttt{v\_proj\_18}    & 0.76 & \texttt{v\_proj\_24}    & 0.80 & \texttt{up\_proj\_17}   & 0.77 \\
    27 & \texttt{v\_proj\_22}    & 0.76 & \texttt{up\_proj\_14}   & 0.78 & \texttt{v\_proj\_25}    & 0.77 \\
    28 & \texttt{v\_proj\_6}     & 0.76 & \texttt{v\_proj\_15}    & 0.78 & \texttt{v\_proj\_31}    & 0.77 \\
    29 & \texttt{v\_proj\_7}     & 0.76 & \texttt{v\_proj\_7}     & 0.78 & \texttt{v\_proj\_7}     & 0.77 \\
    30 & \texttt{down\_proj\_24} & 0.71 & \texttt{up\_proj\_16}   & 0.76 & \texttt{k\_proj\_32}    & 0.75 \\
    \bottomrule
  \end{tabular}
\end{table}

\begin{table}[t]
  \centering
  \footnotesize
  \setlength{\tabcolsep}{4pt}
  \caption{Top-30 layers most frequently kept at high precision in the
    MXFP6/MXFP4 mixed-precision experiments for the Qwen family.
    ``Freq.'' is the fraction of configurations in which the layer was
    assigned the high-precision (MXFP6) format.}
  \label{tab:freq_qwen_mxfp6}
  \begin{tabular}{r ll ll ll}
    \toprule
    & \multicolumn{2}{c}{Qwen3-1.7B}
    & \multicolumn{2}{c}{Qwen3-4B}
    & \multicolumn{2}{c}{Qwen3-8B} \\
    \cmidrule(lr){2-3}\cmidrule(lr){4-5}\cmidrule(lr){6-7}
    Rank & Layer & Freq. & Layer & Freq. & Layer & Freq. \\
    \midrule
    1  & \texttt{down\_proj\_2}  & 1.00 & \texttt{down\_proj\_35} & 0.96 & \texttt{down\_proj\_6}  & 1.00 \\
    2  & \texttt{down\_proj\_27} & 1.00 & \texttt{down\_proj\_6}  & 0.96 & \texttt{v\_proj\_33}    & 0.98 \\
    3  & \texttt{up\_proj\_27}   & 1.00 & \texttt{v\_proj\_33}    & 0.96 & \texttt{down\_proj\_35} & 0.96 \\
    4  & \texttt{v\_proj\_21}    & 1.00 & \texttt{v\_proj\_34}    & 0.96 & \texttt{v\_proj\_34}    & 0.96 \\
    5  & \texttt{v\_proj\_24}    & 1.00 & \texttt{v\_proj\_29}    & 0.94 & \texttt{gate\_proj\_35} & 0.94 \\
    6  & \texttt{v\_proj\_25}    & 1.00 & \texttt{up\_proj\_35}   & 0.92 & \texttt{up\_proj\_35}   & 0.94 \\
    7  & \texttt{v\_proj\_26}    & 1.00 & \texttt{v\_proj\_22}    & 0.92 & \texttt{v\_proj\_22}    & 0.92 \\
    8  & \texttt{v\_proj\_27}    & 1.00 & \texttt{v\_proj\_26}    & 0.92 & \texttt{v\_proj\_27}    & 0.92 \\
    9  & \texttt{v\_proj\_3}     & 1.00 & \texttt{v\_proj\_30}    & 0.92 & \texttt{v\_proj\_30}    & 0.92 \\
    10 & \texttt{v\_proj\_4}     & 1.00 & \texttt{v\_proj\_32}    & 0.92 & \texttt{v\_proj\_26}    & 0.88 \\
    11 & \texttt{v\_proj\_9}     & 1.00 & \texttt{v\_proj\_35}    & 0.92 & \texttt{v\_proj\_29}    & 0.88 \\
    12 & \texttt{v\_proj\_15}    & 0.91 & \texttt{v\_proj\_16}    & 0.89 & \texttt{v\_proj\_32}    & 0.88 \\
    13 & \texttt{v\_proj\_19}    & 0.91 & \texttt{v\_proj\_27}    & 0.89 & \texttt{v\_proj\_35}    & 0.88 \\
    14 & \texttt{v\_proj\_23}    & 0.91 & \texttt{v\_proj\_17}    & 0.87 & \texttt{v\_proj\_16}    & 0.86 \\
    15 & \texttt{v\_proj\_5}     & 0.91 & \texttt{down\_proj\_16} & 0.85 & \texttt{v\_proj\_17}    & 0.86 \\
    16 & \texttt{v\_proj\_8}     & 0.91 & \texttt{gate\_proj\_35} & 0.85 & \texttt{v\_proj\_19}    & 0.86 \\
    17 & \texttt{k\_proj\_3}     & 0.82 & \texttt{v\_proj\_10}    & 0.85 & \texttt{v\_proj\_10}    & 0.84 \\
    18 & \texttt{o\_proj\_0}     & 0.82 & \texttt{v\_proj\_23}    & 0.85 & \texttt{v\_proj\_15}    & 0.84 \\
    19 & \texttt{up\_proj\_11}   & 0.82 & \texttt{v\_proj\_31}    & 0.85 & \texttt{v\_proj\_24}    & 0.84 \\
    20 & \texttt{up\_proj\_26}   & 0.82 & \texttt{v\_proj\_9}     & 0.85 & \texttt{v\_proj\_25}    & 0.84 \\
    21 & \texttt{up\_proj\_5}    & 0.82 & \texttt{k\_proj\_32}    & 0.83 & \texttt{up\_proj\_15}   & 0.82 \\
    22 & \texttt{v\_proj\_16}    & 0.82 & \texttt{v\_proj\_19}    & 0.83 & \texttt{v\_proj\_20}    & 0.82 \\
    23 & \texttt{v\_proj\_17}    & 0.82 & \texttt{v\_proj\_25}    & 0.83 & \texttt{v\_proj\_7}     & 0.82 \\
    24 & \texttt{v\_proj\_18}    & 0.82 & \texttt{k\_proj\_33}    & 0.81 & \texttt{v\_proj\_8}     & 0.82 \\
    25 & \texttt{v\_proj\_20}    & 0.82 & \texttt{k\_proj\_34}    & 0.81 & \texttt{v\_proj\_9}     & 0.82 \\
    26 & \texttt{v\_proj\_7}     & 0.82 & \texttt{v\_proj\_15}    & 0.81 & \texttt{up\_proj\_14}   & 0.80 \\
    27 & \texttt{down\_proj\_19} & 0.73 & \texttt{v\_proj\_20}    & 0.81 & \texttt{up\_proj\_16}   & 0.76 \\
    28 & \texttt{down\_proj\_26} & 0.73 & \texttt{v\_proj\_8}     & 0.81 & \texttt{down\_proj\_16} & 0.75 \\
    29 & \texttt{gate\_proj\_2}  & 0.73 & \texttt{v\_proj\_24}    & 0.77 & \texttt{up\_proj\_13}   & 0.75 \\
    30 & \texttt{gate\_proj\_27} & 0.73 & \texttt{v\_proj\_7}     & 0.77 & \texttt{up\_proj\_18}   & 0.75 \\
    \bottomrule
  \end{tabular}
\end{table}

\begin{table}[t]
  \centering
  \footnotesize
  \setlength{\tabcolsep}{4pt}
  \caption{Top-30 layers most frequently kept at high precision for
    SmolLM2-1.7B, in the MXFP8/MXFP4 and MXFP6/MXFP4 mixed-precision
    experiments. ``Freq.'' is the fraction of configurations in which the
    layer was assigned the high-precision format.}
  \label{tab:freq_smollm2}
  \begin{tabular}{r ll ll}
    \toprule
    & \multicolumn{2}{c}{MXFP8/MXFP4}
    & \multicolumn{2}{c}{MXFP6/MXFP4} \\
    \cmidrule(lr){2-3}\cmidrule(lr){4-5}
    Rank & Layer & Freq. & Layer & Freq. \\
    \midrule
    1  & \texttt{down\_proj\_1}  & 1.00 & \texttt{down\_proj\_1}  & 1.00 \\
    2  & \texttt{down\_proj\_23} & 1.00 & \texttt{down\_proj\_23} & 1.00 \\
    3  & \texttt{down\_proj\_7}  & 1.00 & \texttt{down\_proj\_7}  & 1.00 \\
    4  & \texttt{v\_proj\_2}     & 1.00 & \texttt{up\_proj\_23}   & 1.00 \\
    5  & \texttt{v\_proj\_3}     & 1.00 & \texttt{v\_proj\_2}     & 1.00 \\
    6  & \texttt{up\_proj\_23}   & 0.94 & \texttt{v\_proj\_3}     & 1.00 \\
    7  & \texttt{v\_proj\_4}     & 0.94 & \texttt{v\_proj\_5}     & 1.00 \\
    8  & \texttt{v\_proj\_5}     & 0.94 & \texttt{v\_proj\_6}     & 1.00 \\
    9  & \texttt{v\_proj\_14}    & 0.88 & \texttt{v\_proj\_1}     & 0.91 \\
    10 & \texttt{v\_proj\_15}    & 0.88 & \texttt{v\_proj\_10}    & 0.91 \\
    11 & \texttt{v\_proj\_6}     & 0.88 & \texttt{v\_proj\_12}    & 0.91 \\
    12 & \texttt{v\_proj\_7}     & 0.88 & \texttt{v\_proj\_14}    & 0.91 \\
    13 & \texttt{v\_proj\_8}     & 0.88 & \texttt{v\_proj\_4}     & 0.91 \\
    14 & \texttt{v\_proj\_9}     & 0.88 & \texttt{v\_proj\_7}     & 0.91 \\
    15 & \texttt{gate\_proj\_23} & 0.82 & \texttt{v\_proj\_8}     & 0.91 \\
    16 & \texttt{v\_proj\_10}    & 0.82 & \texttt{v\_proj\_9}     & 0.91 \\
    17 & \texttt{v\_proj\_12}    & 0.82 & \texttt{gate\_proj\_0}  & 0.82 \\
    18 & \texttt{v\_proj\_13}    & 0.82 & \texttt{k\_proj\_0}     & 0.82 \\
    19 & \texttt{v\_proj\_17}    & 0.82 & \texttt{up\_proj\_0}    & 0.82 \\
    20 & \texttt{up\_proj\_0}    & 0.76 & \texttt{v\_proj\_11}    & 0.82 \\
    21 & \texttt{v\_proj\_1}     & 0.76 & \texttt{v\_proj\_13}    & 0.82 \\
    22 & \texttt{v\_proj\_11}    & 0.76 & \texttt{v\_proj\_15}    & 0.82 \\
    23 & \texttt{v\_proj\_16}    & 0.76 & \texttt{v\_proj\_16}    & 0.82 \\
    24 & \texttt{v\_proj\_18}    & 0.76 & \texttt{v\_proj\_17}    & 0.82 \\
    25 & \texttt{v\_proj\_19}    & 0.76 & \texttt{v\_proj\_20}    & 0.82 \\
    26 & \texttt{down\_proj\_0}  & 0.71 & \texttt{down\_proj\_18} & 0.73 \\
    27 & \texttt{gate\_proj\_19} & 0.71 & \texttt{down\_proj\_19} & 0.73 \\
    28 & \texttt{gate\_proj\_22} & 0.71 & \texttt{down\_proj\_20} & 0.73 \\
    29 & \texttt{up\_proj\_18}   & 0.71 & \texttt{gate\_proj\_23} & 0.73 \\
    30 & \texttt{up\_proj\_21}   & 0.71 & \texttt{up\_proj\_16}   & 0.73 \\
    \bottomrule
  \end{tabular}
\end{table}

\subsection{Frequency Heatmap}\label{app:sec:frequency_heatmap}
\begin{figure}[h!]
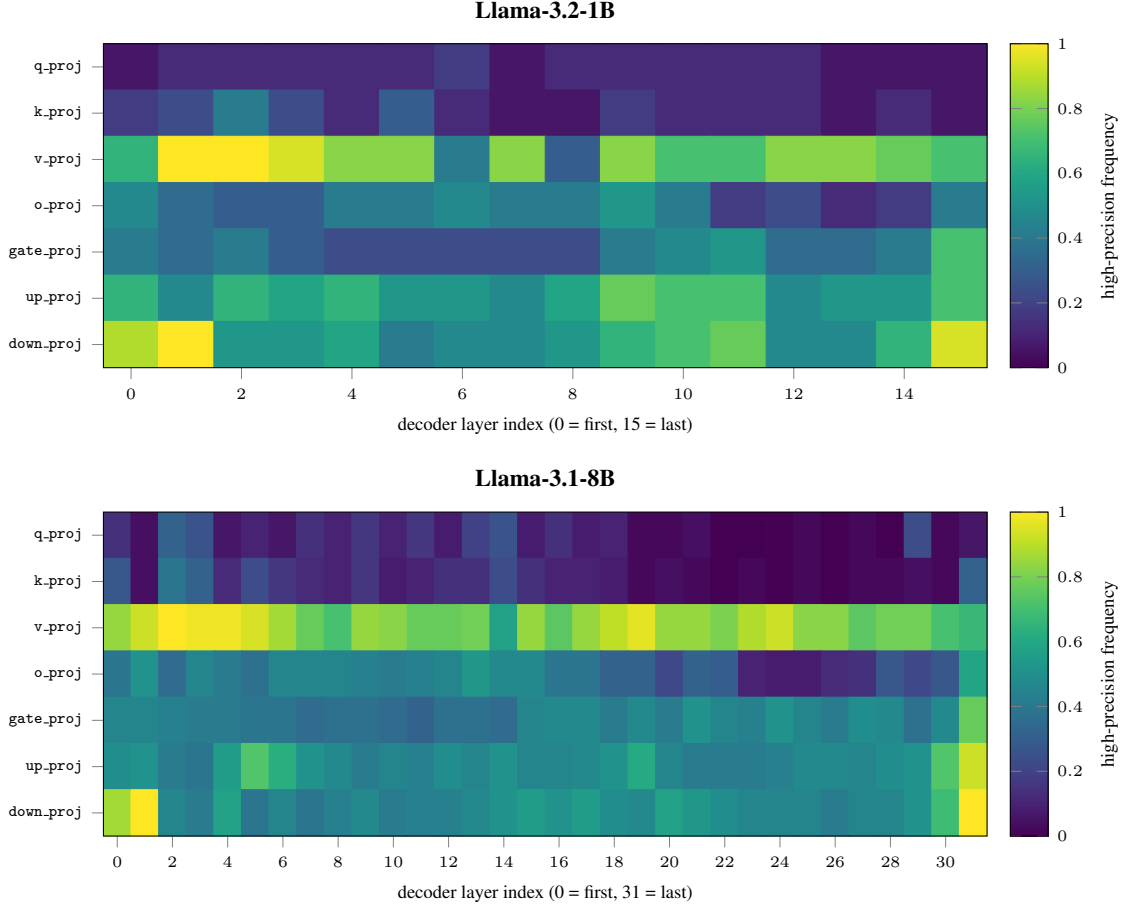

    \centering
    \input{images/frequency_heatmap_llama1b.tex}\\[0.6em]
    \input{images/frequency_heatmap_llama.tex}
    \caption{High-precision layer-frequency heatmaps for two Llama models under
        the MXFP8/MXFP4 mixed-precision configuration:
        Llama-3.2-1B (top, $16$ layers) and Llama-3.1-8B (bottom, $32$ layers).
        Rows: tensor (projection) types;
        columns: decoder layer index ($0$ = first).
        Colour encodes the fraction of configurations in which the tensor is
        assigned the high-precision (MXFP8) format, from $0$ (dark purple,
        always low-precision) to $1$ (yellow, always high-precision).
        In both models the \texttt{v\_proj} and
        \texttt{down\_proj} tensors are most frequently retained at high
        precision, and sensitivity concentrates in the earliest and latest
        decoder layers; the larger model shows the same structure with a
        sharper high/low-precision contrast.}
    \label{fig:freq_heatmap_llama}
\end{figure}

To complement the Tables~\ref{tab:freq_llama_mxfp8}--\ref{tab:freq_smollm2},
\HFigure{fig:freq_heatmap_llama} visualises the complete
per-layer sensitivity maps for two Llama models of different sizes,
Llama-3.2-1B ($16$ layers) and Llama-3.1-8B ($32$ layers),
under the MXFP8/MXFP4 configuration.
Each cell shows the fraction of mixed-precision configurations in which the
corresponding tensor is kept at the high-precision (MXFP8) format:
rows correspond to the tensor (projection) types and columns to the decoder
layer index.

The two heatmaps make the shared pattern visually explicit:
in both models the \texttt{v\_proj} and \texttt{down\_proj} rows are dominated
by high frequencies (yellow), confirming that these tensors are the most
sensitive to quantization regardless of model size, whereas the attention
query/key projections (\texttt{q\_proj}, \texttt{k\_proj}) are almost always
safe to quantize (dark purple).
The effect is also layer-dependent and structurally similar across scales:
sensitivity concentrates in the first few and the last decoder layers, while
the middle of the network tolerates low-precision assignments more readily.
This suggests that the qualitative sensitivity structure transfers across
model sizes even though the exact per-layer ranking does not.

\begin{figure}[h!]
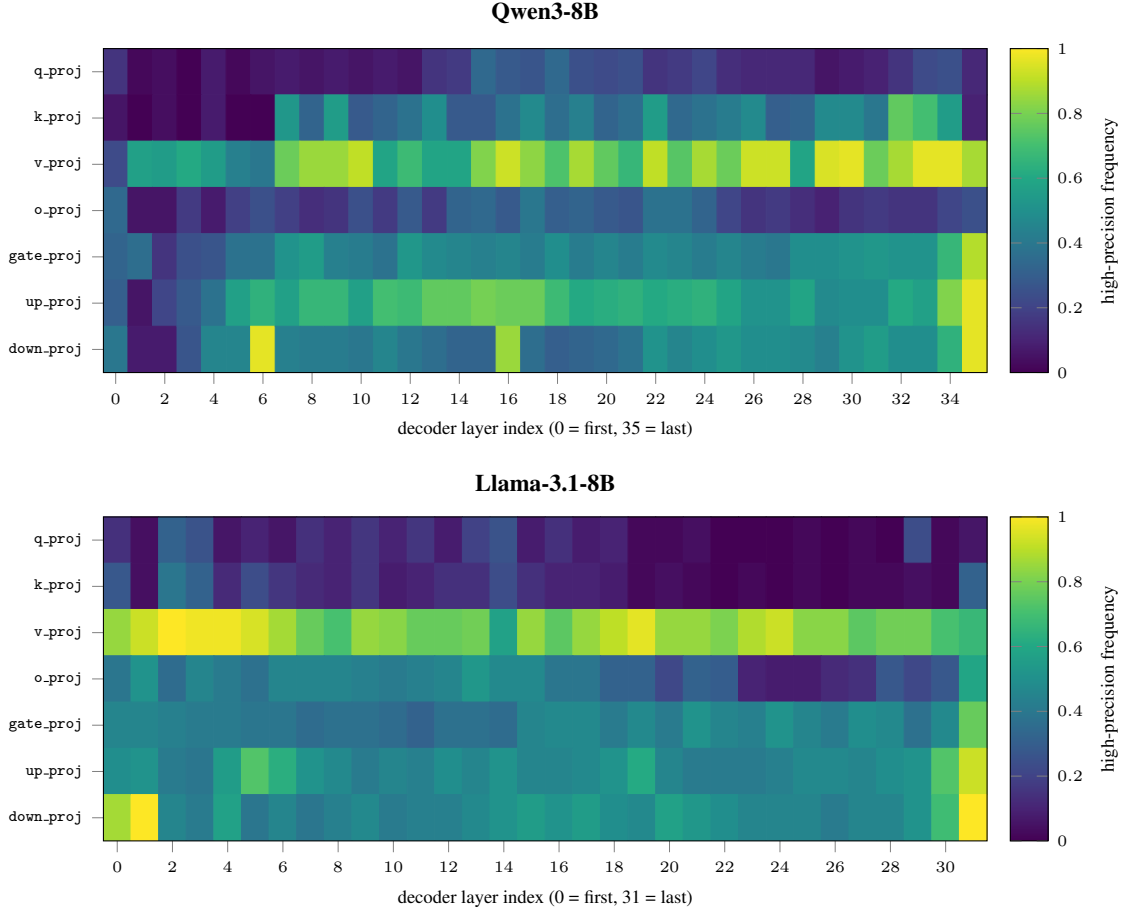

    \centering
    \input{images/frequency_heatmap_qwen8b.tex}\\[0.6em]
    \input{images/frequency_heatmap_llama.tex}
    \caption{High-precision layer-frequency heatmaps for two same-size models
        from different families under the MXFP8/MXFP4 mixed-precision
        configuration:
        Qwen3-8B (top, $36$ layers) and Llama-3.1-8B (bottom, $32$ layers).
        Rows: tensor (projection) types;
        columns: decoder layer index ($0$ = first).
        Colour encodes the fraction of configurations in which the tensor is
        assigned the high-precision (MXFP8) format, from $0$ (dark purple,
        always low-precision) to $1$ (yellow, always high-precision).
        Across both families the \texttt{v\_proj} and
        \texttt{down\_proj} tensors are most frequently kept at high precision,
        although at different indices,
        and the query/key projections are the safest to quantize.
        Qwen3-8B distributes its high-precision assignments
        more broadly across \texttt{gate\_proj}/\texttt{up\_proj} and across
        network depth, while Llama-3.1-8B concentrates sensitivity more sharply
        at the first and last layers, showing that the fine-grained per-layer
        ranking is architecture-specific.}
    \label{fig:freq_heatmap_cross}
\end{figure}

Having established that the sensitivity structure is stable \emph{within} a
model family, we also investigate if this also holds \emph{across} families.
\HFigure{fig:freq_heatmap_cross} compares two models of the same size,
Qwen3-8B and Llama-3.1-8B, under the same MXFP8/MXFP4 configuration.
The comparison reveals that the coarse pattern is largely architecture-agnostic,
although with some notable differences.
In both families the \texttt{v\_proj} and \texttt{down\_proj} tensors again form
the most consistently high-precision rows, and the attention query/key
projections remain the safest to quantize.
However, the two families differ in the details:
Qwen3-8B spreads its high-precision assignments across a broader set of
\texttt{gate\_proj}/\texttt{up\_proj} layers, and its sensitive layers are
distributed more uniformly through the network depth, whereas Llama-3.1-8B
concentrates them more strongly at the boundaries.
This confirms the observation of the previous paragraph that, while the dominant
tensor-type sensitivity transfers across models, the exact per-layer ranking is
architecture-specific and cannot be assumed to carry over between families.

\end{document}